\documentclass[a4paper,fleqn]{cas-dc}
\usepackage[natbibapa]{apacite}
\usepackage{soulutf8}
\usepackage{mathtools, cuted}
\usepackage{algorithm}
\usepackage{algpseudocode}

\usepackage{balance}
\usepackage[subrefformat=parens,labelformat=parens]{subcaption} 
\usepackage{tabularx} 
\usepackage{multirow}
\usepackage{graphicx}
\usepackage[justification=centering]{caption}
\usepackage[subrefformat=parens,labelformat=parens]{subcaption} 

\usepackage{mathtools}

\usepackage{float}
\usepackage{hyperref}
\usepackage{fontawesome5}
\usepackage{geometry
}

\renewcommand{\hl}[1]{#1}

\begin{document}
\begin{titlepage}
    \newgeometry{left=0.5in, right=0.5in}
    \centering
    \vspace*{1cm}
    {\LARGE Dense Neural Network Based Arrhythmia Classification on Low-cost and Low-compute Micro-controller} \\ 
    \vspace*{0.1cm}
    
    {\large \url{https://github.com/MohammedZ666/DenseECGMicro.git}} \\
    
    \vspace*{0.1cm}
    { 
        © 2023. This manuscript version is made available under the CC-BY-NC-ND 4.0 license \url{https://creativecommons.org/licenses/by-nc-nd/4.0/} \\
    }
    \raggedright
    \vspace{2cm}
    \begin{enumerate}
        \item Md Abu Obaida Zishan$^a$ \ $|$ \ Corresponding Author  \ $|$ \ \href{mailto:md.abu.obaida.zishan@g.bracu.ac.bd}{\faEnvelope \ md.abu.obaida.zishan@g.bracu.ac.bd} \ $|$ \ \href{tel:+8801837322980}{\faMobile \ +88.01.837.322.980}
        \item H M Shihab$^{b}$\ $|$ \ \href{mailto:h.m.shihab@g.bracu.ac.bd}{\faEnvelope \ h.m.shihab@g.bracu.ac.bd}
        \item Sabik Sadman Islam$^{b}$\ $|$ \ \href{mailto:sabik.sadman.islam@g.bracu.ac.bd}{\faEnvelope \ sabik.sadman.islam@g.bracu.ac.bd}
        \item Maliha Alam Riya$^{b}$ \ $|$ \ \href{mailto:maliha.alam.riya@g.bracu.ac.bd}{ \faEnvelope \ maliha.alam.riya@g.bracu.ac.bd}
        \item Gazi Mashrur Rahman$^{b}$ \ $|$ \ \href{mailto:gazi.mashrur.rahman@g.bracu.ac.bd}{  \faEnvelope \ gazi.mashrur.rahman@g.bracu.ac.bd}
        \item Jannatun Noor$^{a}$ \ $|$ \ \href{mailto:jannatun.noor@bracu.ac.bd}{\faEnvelope \ jannatun.noor@bracu.ac.bd}
    \end{enumerate}
    \vspace{1cm}
    
    $^{a}$Computing for Sustainability and Social Good (C2SG) Research Group, Department of Computer Science and Engineering, School of Data and Sciences, BRAC University, Dhaka, Bangladesh \\ 
    $^{b}$Department of Computer Science and Engineering, BRAC University, Dhaka, Bangladesh
    \vfill
\end{titlepage}

\let\WriteBookmarks\relax
\def\floatpagepagefraction{1}
\def\textpagefraction{.001}
\shorttitle{Dense Neural Network Based Arrhythmia Classification on Low-cost and Low-compute Micro-controller}
\shortauthors{Zishan et al.}

\title [mode = title]{{Dense Neural Network Based Arrhythmia Classification on Low-cost and Low-compute Micro-controller}}




\author[1]{Md Abu Obaida Zishan}[orcid=0000-0002-6119-5863]
\ead{md.abu.obaida.zishan@g.bracu.ac.bd}

\author[2]{H M Shihab}[orcid=0000-0002-6911-751X]
\ead{h.m.shihab@g.bracu.ac.bd}

\author[2]{Sabik Sadman Islam}[orcid=0000-0002-3096-3874]
\ead{sabik.sadman.islam@g.bracu.ac.bd}

\author[2]{Maliha Alam Riya}[orcid=0000-0003-4151-6356]
\ead{maliha.alam.riya@g.bracu.ac.bd}

\author[2]{Gazi Mashrur Rahman}[orcid=0000-0002-5292-1560]
\ead{gazi.mashrur.rahman@g.bracu.ac.bd}

\author[1]{Jannatun Noor}[orcid=0000-0001-9669-151X]
\ead{jannatun.noor@bracu.ac.bd}

\address[1]{Computing for Sustainability and Social Good (C2SG) Research Group, Department of Computer Science and Engineering, School of Data and Sciences, BRAC University, Dhaka, Bangladesh}
\address[2]{Department of Computer Science and Engineering, BRAC University, Dhaka, Bangladesh}

\cortext[cor3]{Corresponding author}



\begin{abstract}
The electrocardiogram (ECG) monitoring device is an expensive albeit essential device for the treatment and diagnosis of cardiovascular diseases (CVD). The cost of this device typically ranges from \$2000 to \$10000. Several studies have implemented ECG monitoring systems in micro-controller units (MCU) to reduce industrial development costs by up to 20 times. However, to match industry-grade systems and display heartbeats effectively, it is essential to develop an efficient algorithm for detecting arrhythmia (irregular heartbeat). Hence in this study, a dense neural network is developed to detect arrhythmia on the \textit{Arduino Nano}. The Nano consists of the ATMega328 microcontroller with a 16MHz clock, 2KB of SRAM, and 32KB of program memory. Additionally, the \textit{AD8232 SparkFun Single-Lead Heart Rate Monitor} is used as the ECG sensor. The implemented neural network model consists of two layers (excluding the input) with 10 and four neurons respectively with \textit{sigmoid} activation function. However, four approaches are explored to choose the appropriate activation functions. The model has a size of 1.267 KB, achieves an F1 score (macro-average) of 78.3\% for classifying four types of arrhythmia, an accuracy rate of 96.38\%, and requires 0.001314 MOps of floating-point operations (FLOPs). 
\hl{This study has achieved notable advancements compared to recent studies, particularly in terms of neural network metrics and the efficient utilization of computational resources.}
\end{abstract}


\maketitle
\section{Introduction}
\label{sec_introduction}
The cost of treatment for cardiovascular diseases (CVDs)  in developing and developed countries is high. In developing countries, the cost can rise as high as USD 3,842 per patient \citep{kumar} and USD 18,953 in developed countries per patient per year \citep{nichols}. In order to ease, the patients suffering from CVDs it is necessary to develop related biomedical systems that are low-cost, efficient yet accurate. ECG (electrocardiogram) monitoring system is one such system that is expensive and has only a few low-cost alternatives. Conventional ECG monitoring systems cost between \$2000 and \$10,000 \citep{faruk}. Therefore, the cost of any developed low-cost ECG monitoring system must be orders of magnitude cheaper having the same capabilities to make any significant impact in reducing the treatment cost of CVDs globally.

\subsection{Background and Motivation}

17.9 million individuals worldwide, or 32\% of all deaths in 2019, were projected to have been caused by CVDs, according to WHO. 85 percent of these deaths were caused by heart attacks and strokes. This suggests that CVDs account for a sizable share of worldwide mortality. The primary cause of mortality in low and middle-income countries is cardiovascular disease (CVD), contrary to popular belief, which holds that this percentage of deaths only comes from high-income and high-middle-income groups \citep{who}.

\begin{figure}[!ht]
\centering
\includegraphics[width=3in]{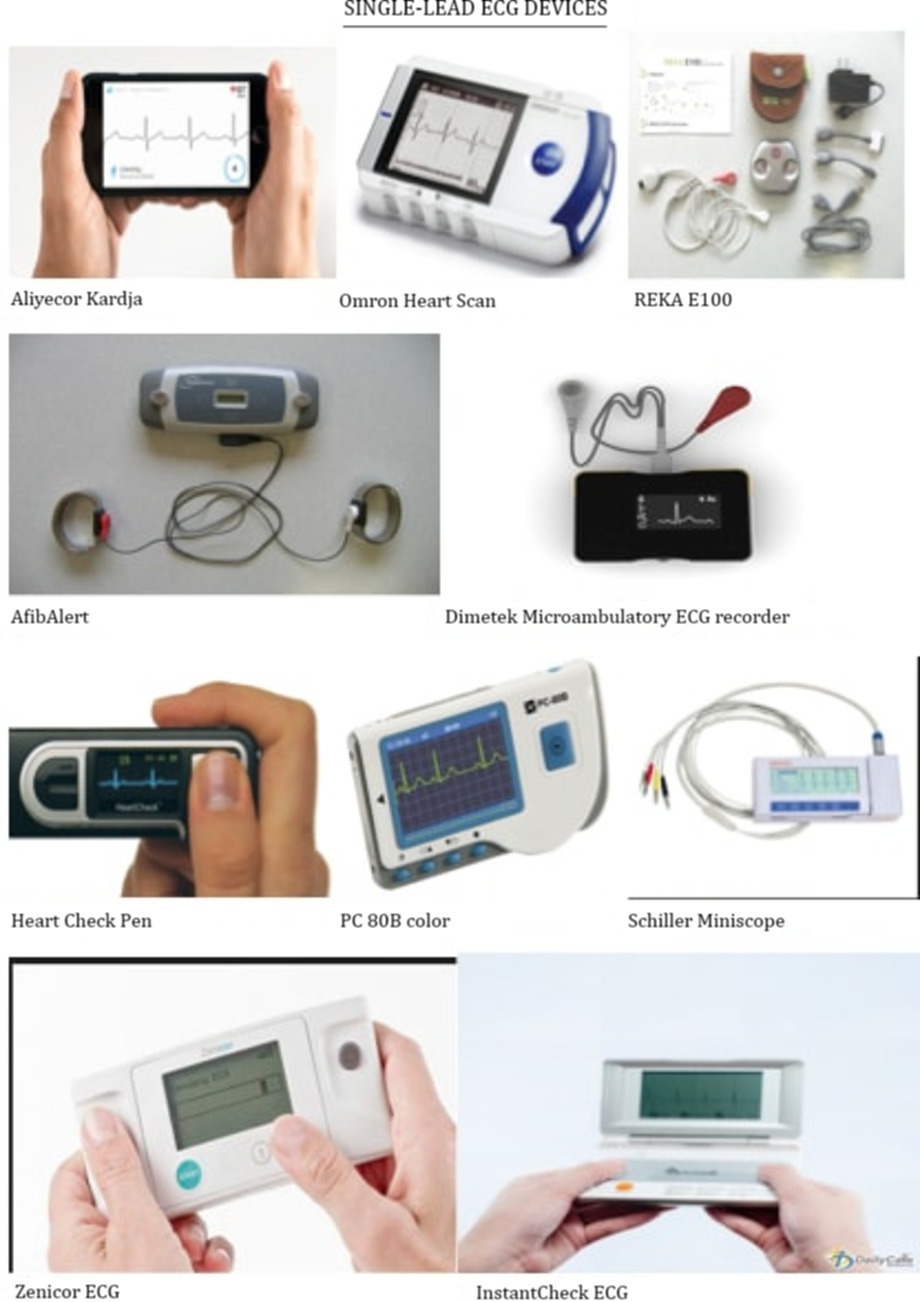}
\caption{Single-lead ECG devices \citep{bansal}}
\label{ecg_devices}
\end{figure}

The ECG, or electrocardiogram machine, is one of the most important tools used to diagnose such disorders. Electrocardiography is simply graphing of a heart's electrical activity \citep{bunce}. A voltage versus time graph of the electrical activity of the heart is called an electrocardiogram \citep{lilly}. Using probes linked to the patient's torso, the ECG monitoring system draws the heartbeat as a voltage vs time graph on a monitor. This allows physicians to assess the patient's pulse patterns. When a patient experiences an arrhythmia or an abnormal heartbeat, modern ECG devices set off alerts. Figure \mbox{\ref{ecg_devices}} shows commercial single-lead ECG devices.

\subsection{Existing Works on Low-cost ECG Monitoring System}
Although recent studies provide successful approaches for developing ECG monitoring systems, not all of them are cost-effective.  Since any ECG monitoring system would require running 24/7, power consumption is also a factor overlooked by most studies. Moreover, ECG monitoring systems are used once for every required patient in a hospital, which further multiplies the importance of low-power and low-cost systems. Chen et al. \citep{chen}, Sakib et al. \citep{sakib}, Ahsanuzzaman et al. \citep{ahsanuzzaman} and Hartman et al. \citep{hartmann} developed an ECG monitoring system on single board computers (SBC). Faraone et al. \citep{faraone}, Scirè et al. \citep{scire} and  Raj et al. \citep{raj} developed the same but with mere micro-controller units (MCUs). Additionally, all of them implemented machine-learning-based arrhythmia detection on their devices including studies where MCUs were used as the main device. 

Using MCUs instead of SBCs lowers the power consumption by a factor of 1000 at the very least since MCUs operate within milliwatt ranges. Moreover, MCUs are comparatively cheaper than SBCs. Hence, for broad-scale deployment and 24/7 duty, MCUs are a much better choice than SBCs for ECG monitoring system development. However, development with MCUs does not necessarily mean sacrificing any features available in modern ECG monitoring systems since some of the above-mentioned studies already proved that.  

Still, one might assume that arrhythmia detection on ECG monitoring systems may require cycle expensive algorithm like machine learning which might not be feasible for MCUs. But inference or forward pass of a small enough machine learning model is possible. Since they are primarily simple matrix multiplications whose values are passed through activation functions of a few varieties. Even, training machine learning models have been possible in  MCUs \citep{lin}. Therefore, with an efficient enough machine learning model for arrhythmia detection, one can develop a cost, compute, and power-efficient ECG monitoring system based on MCUs with little to no drawbacks compared to industry-grade systems.  

Moreover, the high expense of CVD treatment clearly reflects the expensive biomedical equipment (like the ECG) required to treat them. This shows that a cheaper device will lower the treatment cost of CVDs. Developing and undeveloped countries would benefit significantly from affordable CVD treatment since they are currently expensive and the recovery rate is slow. This means an affordable ECG monitoring system would make treatment accessible for more people, from all walks of life, and allow them to get better treatment, saving millions. 

\subsection{Our Contributions}
Therefore, we require an affordable ECG monitoring system without trading away modern features like real-time arrhythmia classification. This is why we make the following contributions to this study --
\begin{itemize}
    \item We perform a comparative study on recent papers on their approach to developing an ECG system. Moreover, we analyze the papers on four metrics -- cost, power, compute, and the use of detection algorithms.
    \item Subsequently, we propose an implementation of a densely connected neural network to detect arrhythmia on a low-compute device. We formulate four such neural networks namely -- Sigmoid-sigmoid, Sigmoid-softmax, ReLU-sigmoid, and ReLU-softmax,  and compare their performance.
    \item Consequently, we implement a heartbeat detection algorithm and a hardware system to ensure the practical applicability of our network in an end-to-end ECG monitoring system.
    \item Finally, we discuss and compare results with the studies consisting of the best approaches from the comparative analysis. Our systems have outperformed those studies in most scales of comparison.

\end{itemize}

\subsection{Organization of This Study}

First, we present background information on ECG monitoring systems in Section \ref{sec_ecg_over}. After, in Section \ref{sec_comp_analysis}, we conduct a comparative analysis of some studies conducted from 2018 to 2021 and show the best approach. Besides, in Section \ref{sec_system_design}, we provide our system's hardware and software design. Next, in Section \ref{sec_adm}, we present our densely connected neural network to detect arrhythmia. Then, we discuss the training setup, results, discussion, and, findings of our implemented system in Section \ref{sec_exp}. Finally, we conclude this paper (in Section \ref{conclusion}) by discussing future research directions and their impact.

\section{Background}
\label{sec_ecg_over}
\begin{figure*} 
\centering
\includegraphics[width=6in]{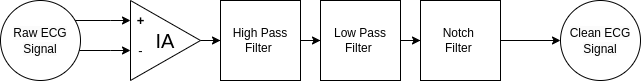}
\caption{Flowchart for ECG amplification and filtration circuitry}
\label{ecg_signal_chain}
\end{figure*}

\begin{figure}[ht]
\centering \includegraphics[width=\columnwidth]{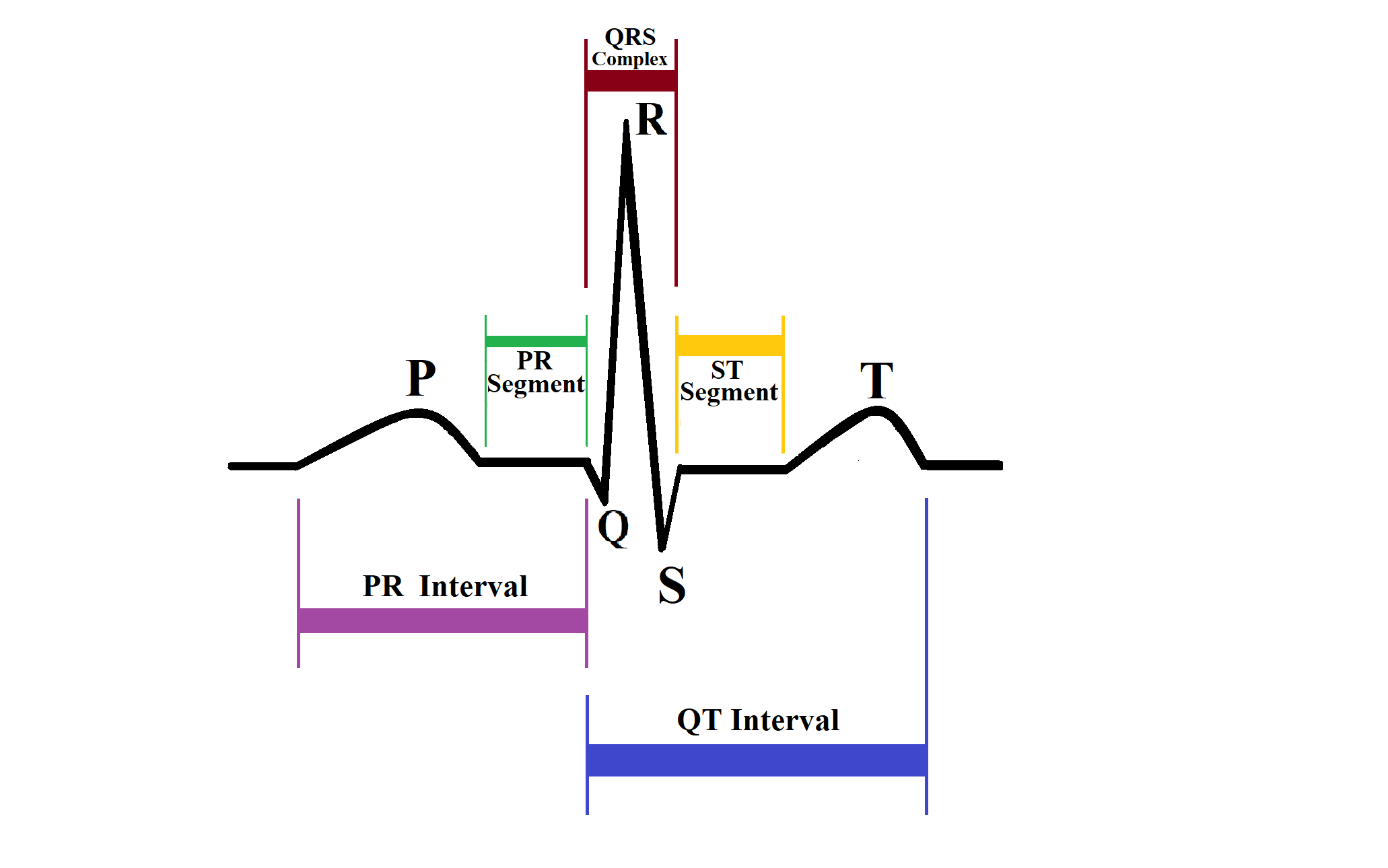}
 \caption{A voltage vs time representation of a heartbeat / QRS complex}
 \label{qrs_signal}
\end{figure}

An ECG monitoring system can be broken down into three components mentioned below --  
\begin{itemize}
    
    \item \textit{Electrodes}: To read the potential difference created by the depolarization and repolarization of the heart, electrodes need to be placed in the limbs and torso of the subject, which generates pure and unfiltered ECG signals. An ECG system may contain single or multiple pairs of electrodes or leads \citep{faruk}. 
    
    \item \textit{Amplification and filtration circuitry}: An ECG signal is within a very low voltage between 0.5 and 4mV range \citep{webster}. It is not easy to read these signals without the electromagnetic interference of much stronger signals like infrastructural AC electricity. Additionally, ECG signals contain noise which must be filtered to read them. Hence, amplification and filtration circuitry (bandpass filter) must be implemented.
    \item  \textit{Heartbeat Detection Algorithms}: Following cleaning and amplification of the ECG signal (as in Figure \ref{ecg_signal_chain}), it has to be passed through a heartbeat detection algorithm to separate heartbeat (also defined as the QRS complex in Figure \ref{qrs_signal}) from the signal. One such algorithm is the Pan-Tomkins algorithm \citep{pan} which uses programmatic bandpass filtering, differentiation, squaring, moving window integration (convolution), and a dual-thresholding technique to detect regions of heartbeats from an ECG signal stream.
    \item \textit{Arrhythmia Detection Algorithm}: After successfully detecting QRS complexes, the advanced ECG systems often detect anomalous heartbeats, also known as arrhythmia to alert the physicians about the probable deterioration of the patient's heart condition.
\end{itemize}

In our study, we develop a novel arrhythmia detection algorithm via dense neural networks so that it can be made efficient enough to fit inside a very low-memory MCU. Moreover, we implement the other three components, to properly simulate real-time detection of arrhythmia and make the arrhythmia detection algorithm compatible and consistent within an ECG monitoring system.

\section{Literature Review and Comparative Study}
\label{sec_comp_analysis}
In this Section, we conduct a brief comparative analysis of studies that tried to implement an ECG system based on four metrics -- a) cost, b) compute c) power, and d) use of arrhythmia detection algorithms. However, the analysis and comparison were done in detail in our previous study \citep{lowecg}. Besides, Chen et al. \citep{chen} developed an ECG system for real-time analysis of atrial fibrillation (AF). They used a PYNQ-Z2 board with a programmable chip and FPGA. For the AF detection algorithm, they used neural networks.  

\begin{table*}[!ht]
\centering
\caption{Comparative analysis based on cost, compute, power and detection algorithm}
\label{comparative_analysis}
\begin{tabular}{|c|c|c|c|c|}
    \hline
    \text{Study}  & \text{Cost (USD)} & \text{Compute (MHz)} & \text{Power (mW)} & \text{Detection Algorithm} \\
    \hline
    Chen et al., 2021  \citep{chen} & 149 & 1300 & - & NN \\
    \hline
    Sakib et al., 2021  \citep{sakib} & 59-99 & 5720 & $\geq$ 700 &CNN\\
    \hline
    Ahsanuzzaman et al., 2020 \citep{ahsanuzzaman} & 35&4800& $\geq$ 700 &LSTM\\
    \hline
    Faraone et al., 2020 \citep{faraone} &39 &64 & 20.65 &R-CNN\\
    \hline
    Hartman et al., 2019 \citep{hartmann} &35 & $\geq$ 700 &$\geq$ 700 &DistilledDNN\\
    \hline
    Scirè et al., 2019  \citep{scire}&39.95&32& 44.08 &KNN + LSTM\\
    \hline
     Our system & 24.90 & 16 & $\sim$ 96 & NN\\
     \hline
\end{tabular}
\end{table*}

Moreover, Sakib et. al \citep{sakib} used both Raspberry Pi 3, 4, and Jetson Nano board as their ECG system's hardware platform. However, the Jetson Nano board had superior performance. They implemented Convolutional Neural Network (CNN) on a single-channel ECG stream to detect arrhythmia. In another study, Ahsanuzzaman et al. and Hartmann et al. \citep{ahsanuzzaman, hartmann} used Raspberry Pi as their main development board for their ECG system. Ahsanuzzaman et al. used a Long-Short Term Memory (LSTM) algorithm for arrhythmia detection whereas Hartmann et al. used distilled deep neural networks (dDNN).

Furthermore, Faraone et al. \citep{faraone} implemented a convolutional recurrent neural network (RCNN) for their arrhythmia detection algorithm. They used the nRF52 chip for the implementation of their system. Scirè et al. \citep{scire} developed a heartbeat detection algorithm using k-Nearest Neighbour (KNN) and LSTM for arrhythmia classification. For their implementation hardware, they used the Intel Curie module and NXP MPC8572E module.

We observe from Table \ref{comparative_analysis} that Scirè et al. and Faraone et al. had orders of magnitude less amounts of power and compute consumed (44.08 mW and 20.65 mW respectively). In contrast, a 700 MHz or more clocked SBC running a Linux operating system could draw at least 700mW at idle mode with no peripherals. This shows that it is possible to run a machine learning-based arrhythmia classification/detection algorithm in a very low-computing system as a microcontroller unit (MCU).

\section{Proposed Methodology and System Design}
\label{sec_system_design}
In this Section, we provide our own implementation of an ECG system. The outline of the entire system and the compute, cost, and power parameters are discussed as follows:

\begin{figure}[h]
\centering
\includegraphics[width=\columnwidth]{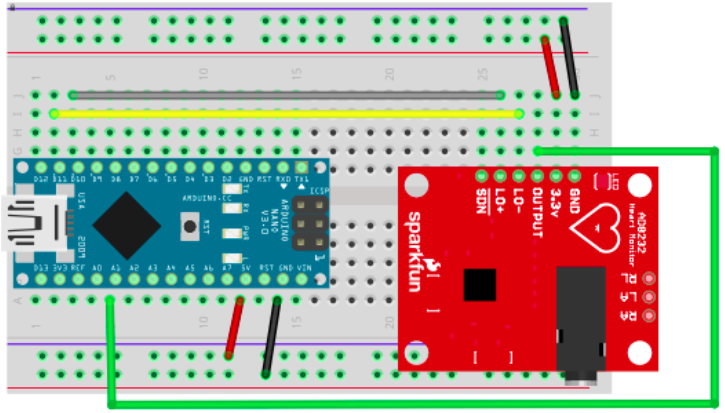}
\caption{System hardware setup}
\label{our_setup}
\end{figure}

\subsection{Hardware} 
For our development board we chose Arduino Nano. Additionally, we used the AD8232 SparkFun Single-Lead Heart Rate Monitor sensor to read the ECG data stream. Figure \mbox{\ref{our_setup}} shows our hardware setup. The details are as follows:

\subsubsection{Arduino Nano}
\label{nano_discuss}
We chose the Arduino Nano \mbox{\citep{nano}} as our microcontroller over its competitions for the following reasons: 

\begin{itemize}
    \item \textit{Popularity:} The Arduino Nano is still the most popular MCU for both hobbyists and professionals \citep{arduino_popular}. Moreover, the documentation regarding the nano is well-written, and much help can be obtained from popular online forums like Stack Overflow \mbox{\citep{stackoverflow}}. Hence, using the Arduino Nano for our system’s implementation ensures easy reproducibility, experimentation, and future development of our system for beginners and experts alike.
    \item \textit{Low compute:} The Arduino Nano has the lowest specification in terms of SRAM, clock speed, and program memory, compared to its recent competitors like the RaspeberryPi PICO, ESP32, and STM32 (Blue and Black Pill). Hence, if our model can be constrained to run on the Nano, it will run on other MCUs with higher specifications as well. However, this is not necessarily true vice versa.
    \item \textit{Low power:} The Arduino Nano consumes low power of up to 95 mW compared to a lot of its competitors. This eventually reduces the electricity bill for running numerous ECG systems in a hospital setup.
    \item \textit{Low-cost:} Even though the cost of the Arduino Nano has increased recently, it is still cheaper than most single-board computers and MCUs. Therefore, low cost is certainly one of the factors for choosing this development board for our experimentation.
\end{itemize}

Table {\ref{nano_spec}} compares the Arduino Nano with its more recent competitors with their hardware specifications.

\begin{table*}[!ht]
\centering
\caption{Specifications of the Arduino Nano, Raspberry Pi PICO, ESP32 and STM32 based microcontroller boards}
\label{nano_spec}
\begin{tabular} {|c|c|c|c|c|}
\hline
\multicolumn{1}{|p{3cm}}{\centering Boards} & \multicolumn{1}{|p{3cm}}{\centering Arduino Nano {\citep{nano}}}& \multicolumn{1}{|p{3cm}}{\centering Raspberry Pi PICO {\citep{pico}}} & \multicolumn{1}{|p{3cm}}{\centering ESP32 {\citep{esp32}}} & \multicolumn{1}{|p{3cm}|}{\centering  STM32 Blue and Black Pill {\citep{stm32black,stm32blue}}} \\
\hline
\multicolumn{1}{|p{3cm}}{\centering Processor} & \multicolumn{1}{|p{3cm}}{\centering ATMega328} & \multicolumn{1}{|p{3cm}}{\centering Dual ARM 32-bit Cortex M0+} & \multicolumn{1}{|p{3cm}}{\centering Xtensa single/dual-core 32-bit LX6 microprocessor(s)}& \multicolumn{1}{|p{3cm}|}{\centering ARM 32-bit Cortex M3/M4} \\
\hline
\text{Compute (MHz)} &16& 133 & 80-240 & 72-84\\
\hline
\text{SRAM (KB)} &2 & 264 & 520 & 20-64\\
\hline
\text{Flash memory (KB)} &32& 2048 & 448 & 64-256 \\
\hline    
\text{Power (mW)} & 95 & 428 - 464 & 66-224.4 & 81-1000\\
\hline
\text{Cost (USD)} &24.90 & 4 & 5-10 & 5-10\\
\hline 
\end{tabular}
\end{table*}

\subsubsection{AD8232 SparkFun Single-Lead Heart Rate Monitor}
The AD8232 SparkFun Single-Lead Heart Rate Monitor is a low-cost board for measuring the electrical activity of the heart. This electrical activity is recorded as an ECG or electrocardiogram and output as an analog measurement. The AD8232 single-channel heart rate monitor acts as an operational amplifier to easily obtain clear signals from the PR and QT intervals  \citep{ecg_sensor}.

\begin{table}[!ht]
\centering
\caption{Sparkfun ECG AD8232 Sensor board  \citep{ecg_sensor}}
\label{sensor_spec}
\begin{tabular} {|c|c|}
\hline
\text{Cost} & USD 21.50  \\
\hline
\text{Leads} & Single\\
\hline
\text{Operating voltage} &  3.3 V\\
\hline
\text{Current consumption} & 170 $\mu$ A\\
\hline
\text{Power} & 0.561 mW \\
\hline
\end{tabular}
\end{table}

\subsection{Software} There are two steps to detecting Arrhythmia from raw ECG data - 
\label{modules_sub}
\begin{itemize}
    \item \textit{Heartbeat Detection Module (HDM)}: The heartbeat detection module detects R peaks from an ECG data stream. We chose the Pan-Tompkins  (PT) algorithm  \citep{pan} which is highly respected and established in the bibliography. However, we chose a modified variant of the algorithm for efficient heartbeat detection \citep{michal}.  
    
    \item \textit{Arrhythmia Detection Module (ADM)}:  After detecting the heartbeat, we need to classify Arrhythmia into four classes only one of which is normal via a densely connected neural network. The arrhythmia detection module is our principal contribution to this study.

\end{itemize}

\begin{figure*}
\centering
\includegraphics[width=0.7\textwidth]{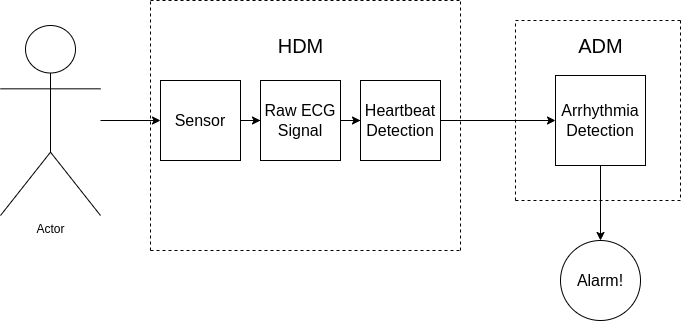}
\caption{System flow-diagram of our ECG monitoring system}
\label{system_flow_diag}
\end{figure*}

In Figure \ref{system_flow_diag}, we present the end-to-end flow diagram of our system design. Raw ECG signal passes from the actor to the processing unit, where the signal is preprocessed. Afterward, the preprocessed signal is passed to the heart-beat detection module (HDM). If the signal is not a heartbeat the algorithm continues otherwise it passes the stream to the arrhythmia detection module (ADM) which if detects an abnormal beat, sets off the on-device alarm. Here, the simple alerting system is only for demonstration, a much more sophisticated alert system can be implemented if needed.

We can see from Table \mbox{\ref{nano_spec}} that power consumption of Nano is only 95mW. However, adding peripherals that draw power may increase power consumption. Table \ref{sensor_spec} depicts the specifications of the Sparkfun ECG AD8232 Sensor board we used. We can see that the power consumption by the board is close to 0.561 mW which barely affects the total power consumption of our system when connected.


\section{Densely Connected Neural Network Based Arrhythmia Detection Module}
\label{sec_adm}
The arrhythmia detection module reads a detected heartbeat from the HDM module and classifies heartbeats into Normal, Supraventricular, Ventricular, and Fusion beats (please refer to Table \ref{arrhythmia_classification}).  For classification, we used dense neural networks.  In this Section, we discuss the data preprocessing, architecture, and, training methodology of our neural network which is the core of our arrhythmia detection module (ADM). 

\begin{table}
\centering
\caption{Classification of Arrhythmia according to AAMI  \citep{aami} }
\label{arrhythmia_classification}
\begin{tabular} {|c|c|}
\hline    
N  & Normal and bundle branch block beat\\
\hline 
S & Supraventricular ectopic beats \\
\hline    
V & Ventricular ectopic beats\\
\hline    
F & Fusion of N and V\\
\hline
Q & undefined or paced beats \\
\hline
\end{tabular}
\end{table}

\subsection{Data Preprocessing}
For training our NN model we used the MIT-BIH dataset \citep{mitbih_dataset}. The Association for the Advancement of Medical Instrumentation (AAMI) standard for testing and reporting performance results of cardiac rhythm and segment measurement algorithms states that the arrhythmia classification performance should be based on classifying five major categories of heartbeats (Table \ref{arrhythmia_classification}). However, we have not classified undefined or paced beats (Q) since these beats are regulated by a pacemaker. Hence, they are not significantly meaningful for arrhythmia classification \mbox{\citep{scire}}.

\begin{figure*}
\centering
\begin{subfigure} {0.49\textwidth}
\includegraphics [width=\textwidth] {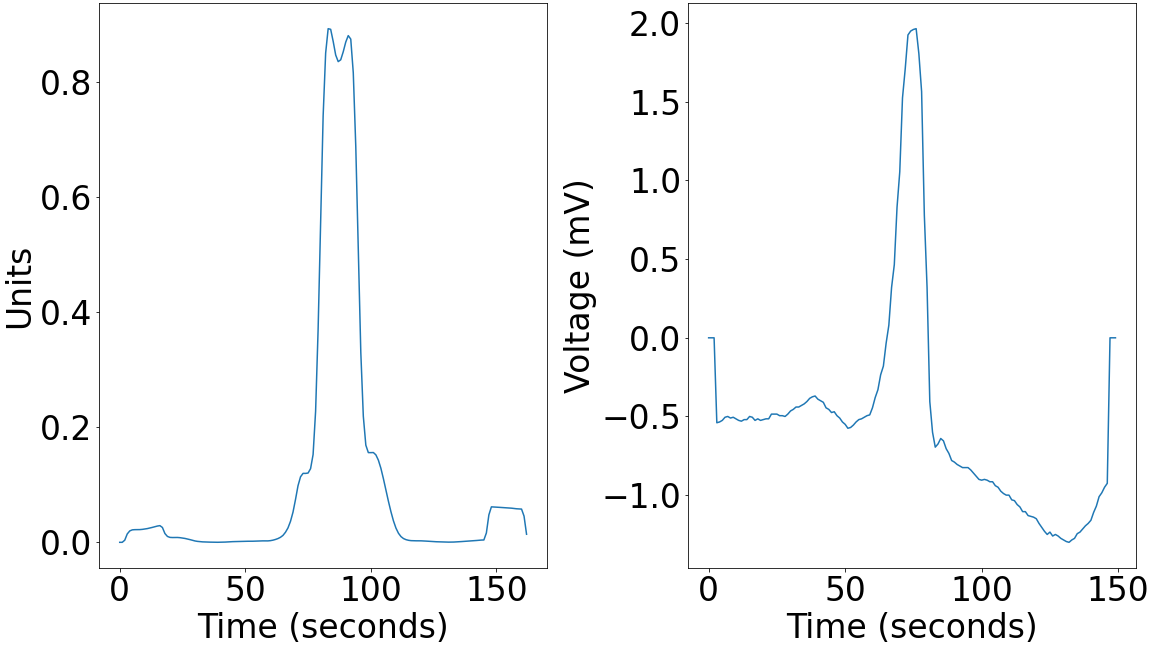}
\caption{Normal beat preprocessed and unchanged}
\label{N_beat}
\end{subfigure}
\begin{subfigure}  {0.49\textwidth}
\includegraphics[width=\textwidth]{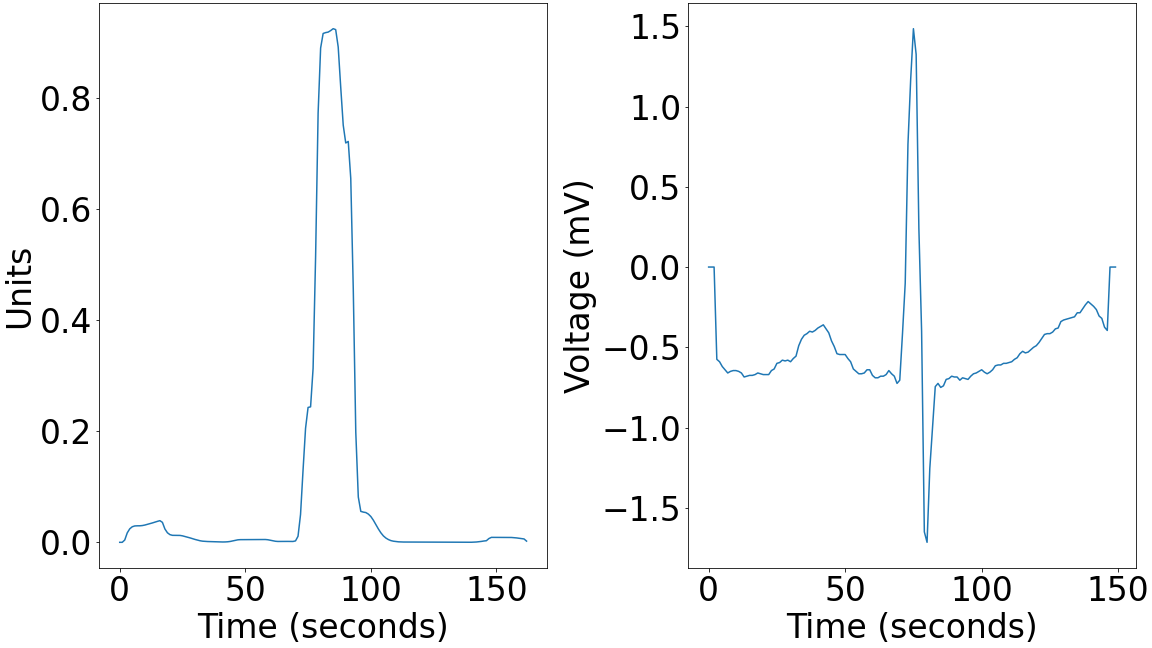}
\caption{Supraventricular ectopic beat preprocessed and unchanged}
\label{S_beat}
\end{subfigure}
\begin{subfigure} {0.49\textwidth}
\includegraphics[width=\textwidth]{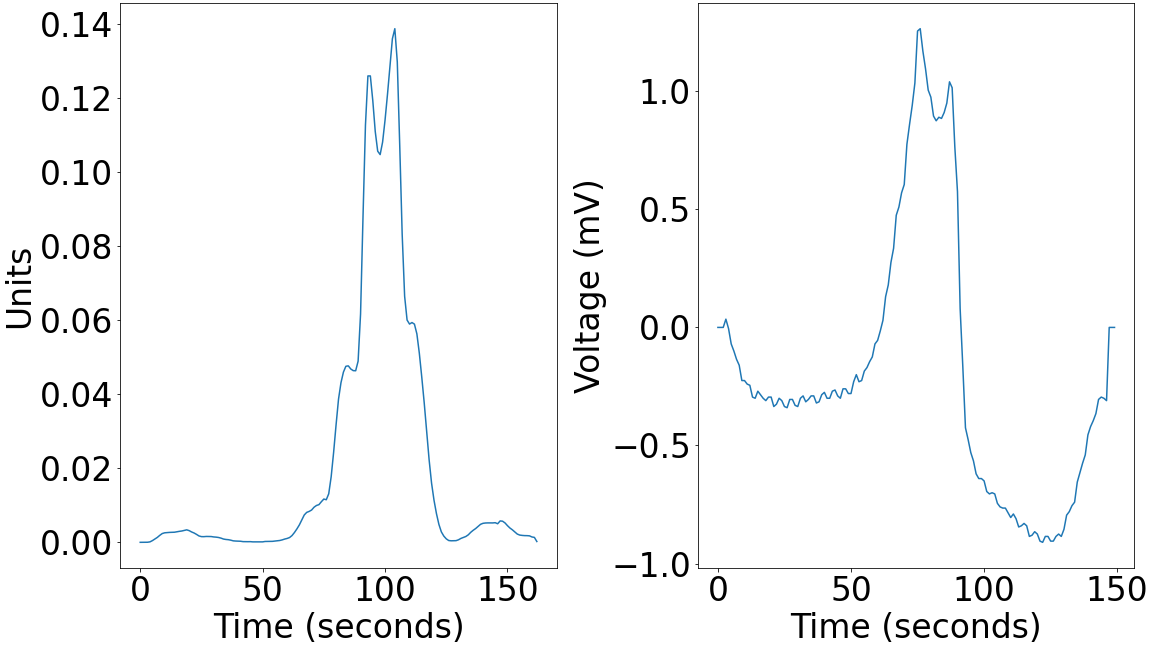}
\caption{Ventricular ectopic beat preprocessed and unchanged}
\label{V_beat}
\end{subfigure}
\begin{subfigure}  {0.49\textwidth}
\includegraphics[width=\textwidth]{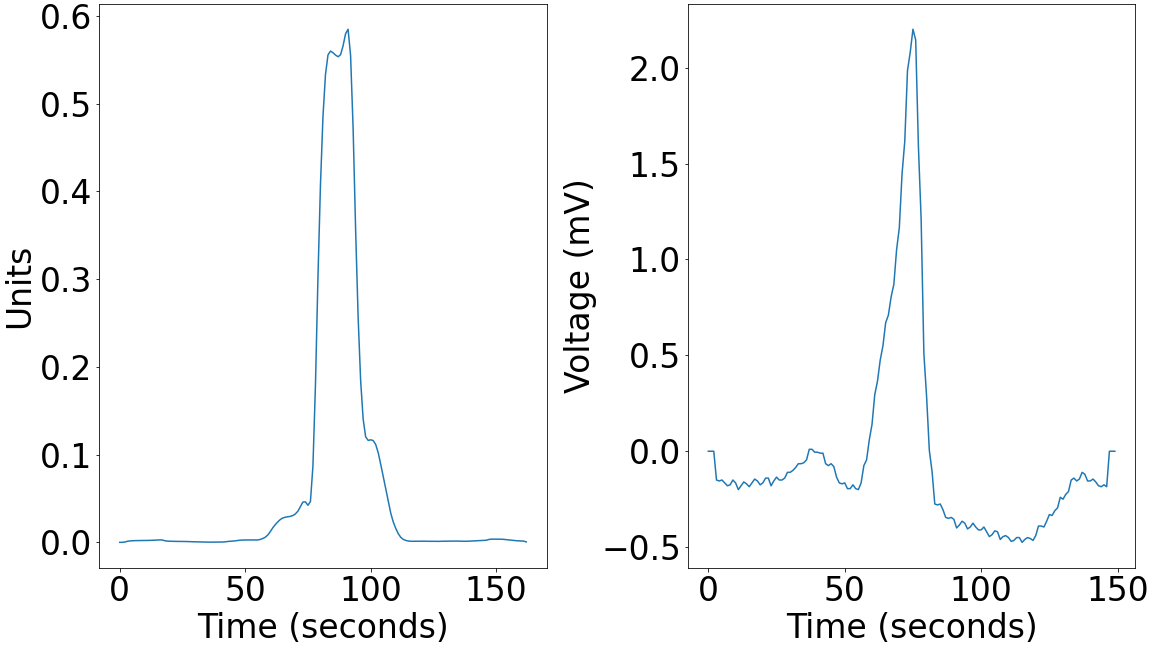}
\caption{Fusion beat preprocessed and unchanged}
\label{F_beat}
\end{subfigure}
\caption{Preprocessed vs unchanged beats of type N, S, V, and F}
\label{fig_beat}
\end{figure*}

In Figure \mbox{\ref{fig_beat}}, we can observe the difference between the preprocessed and unchanged heartbeats consisting of four types of arrhythmia.  The preprocessing follows the same preprocessing technique as the Pan-Tomkins algorithm \mbox{\citep{pan}}. The preprocessing includes a sequence of operations namely - bandpass filtering, differentiation, squaring, and moving window integration. This converges the signal into a smaller temporal space and amplifies the R peak value (maximum value of a heartbeat/QRS complex segment). Since the ECG signal passes through the Pan-Tomkins algorithm for heartbeat detection in the HDM module (discussed in Subsection \mbox{\ref{modules_sub}}) before passing through the neural network, it is necessary that we train our network by applying the same exact sequence of preprocessing techniques. We formulate the preprocessing algorithm in Algorithm \mbox{\ref{algo_preprocess}}

Please note that in Figure \ref{fig_beat}, the y-axis of the preprocessed subfigures (on the left) is written \textit{units}. Due to the transformations, we applied before, we simplify the transformed sensor output from voltage to units. 

\begin{algorithm*}
\caption{Algorithm for training DNN for classification of arrhythmia}
\label{algo_preprocess}
\begin{algorithmic}
\Ensure  $data$ from MIT - BIH dataset is initialized
\State $preprocessed\_beats \gets \emptyset$
\While{$data.length > 0$}
\State $beat, type \gets data.get()$
\State $beat \gets bandpass\_filter(beat \gets beat, high \gets 15 Hz, low \gets 5 Hz)$
\State $beat \gets derivate(beat \gets beat)$
\State $beat \gets square(beat \gets beat)$
\State $beat \gets moving\_window\_integration(beat \gets beat, window \gets 15)$
\State $beat \gets beat.get\_values(start \gets index - 30, end \gets index + 30)$
\State $preprocessed\_beats.append((beat, type))$    
\EndWhile
\State $train\_and\_validate(preprocessed\_beats)$ 
\end{algorithmic}
\end{algorithm*}

\subsection{Implemented Neural Networks}

\begin{figure*}
\centering
\begin{subfigure}[b]{0.33\textwidth}
\centering
\includegraphics[width=\textwidth]{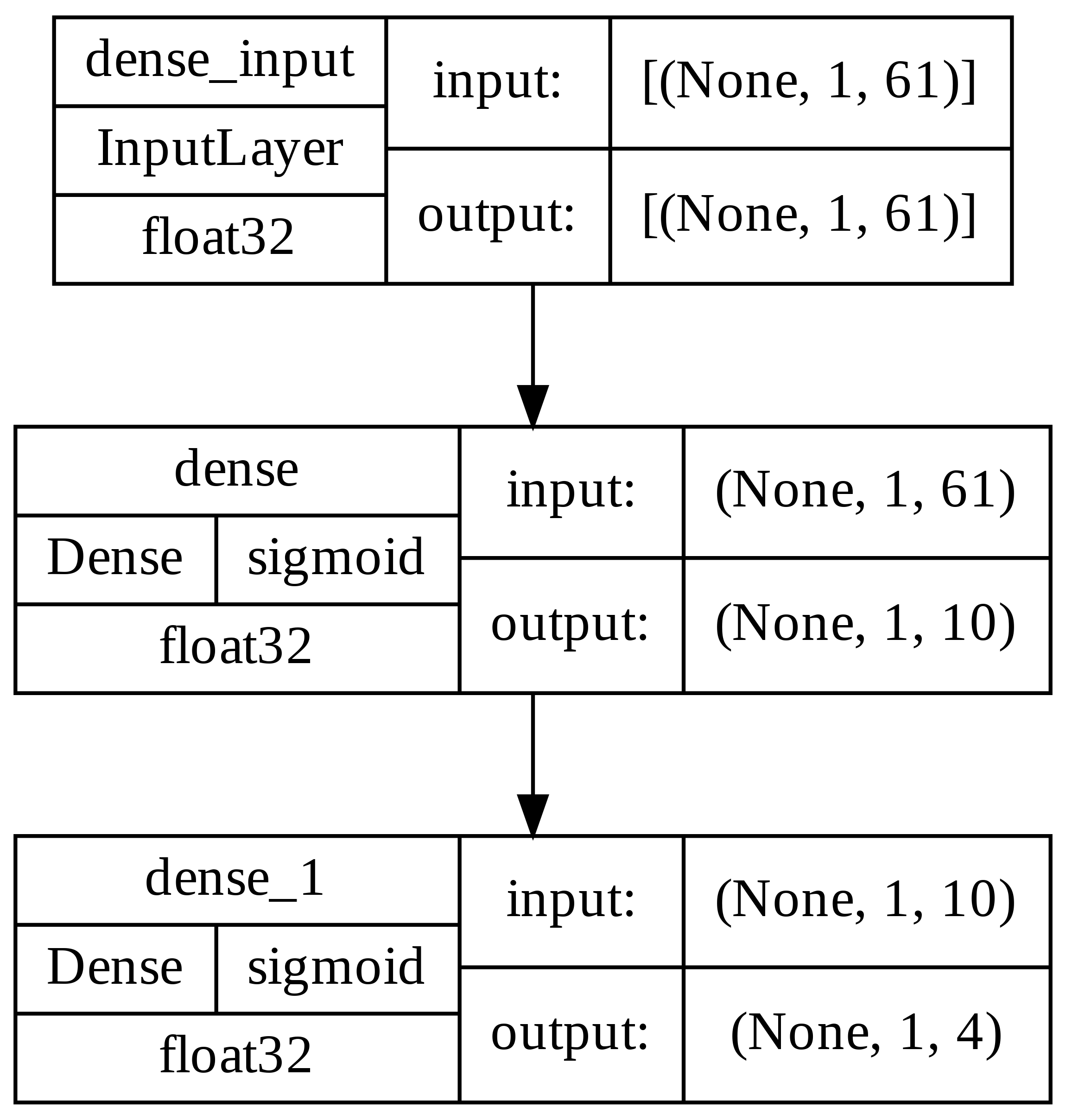}
\caption{Sigmoid-sigmoid based network architecture}
\label{sigmoid_sigmoid_sturcture}
\end{subfigure}
\begin{subfigure}[b]{0.33\textwidth}
\centering
\includegraphics[width=\textwidth]{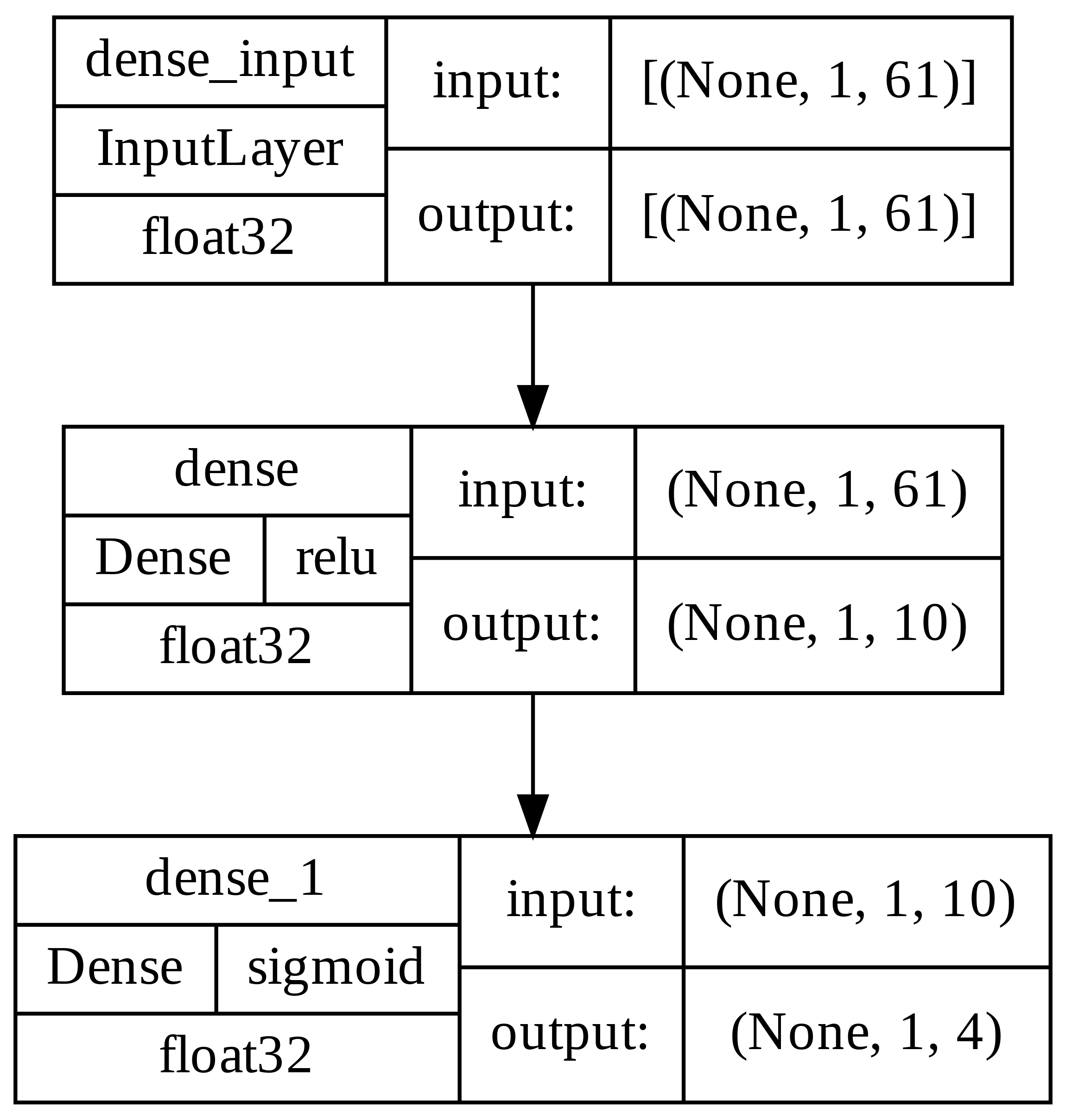}
\caption{ReLU-sigmoid based network architecture}
\label{relu_sigmoid_structure}
\end{subfigure}
\begin{subfigure}[b]{0.33\textwidth}
\centering
\includegraphics[width=\textwidth]{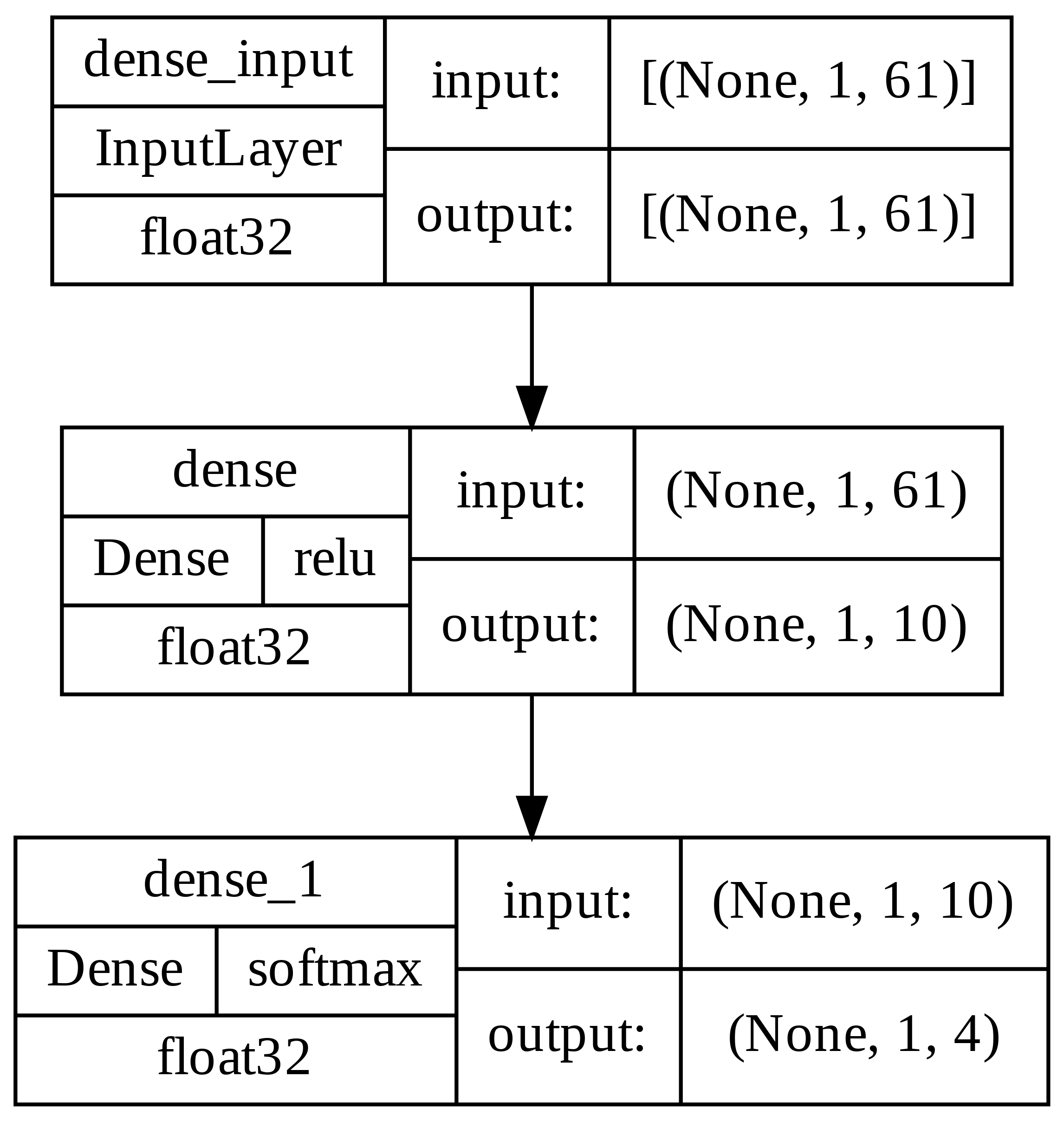}
\caption{ReLU-softmax based network architecture}
\label{relu_softmax_structure}
\end{subfigure}
\begin{subfigure}[b] {0.33\textwidth}
\centering
\includegraphics[width=\textwidth]{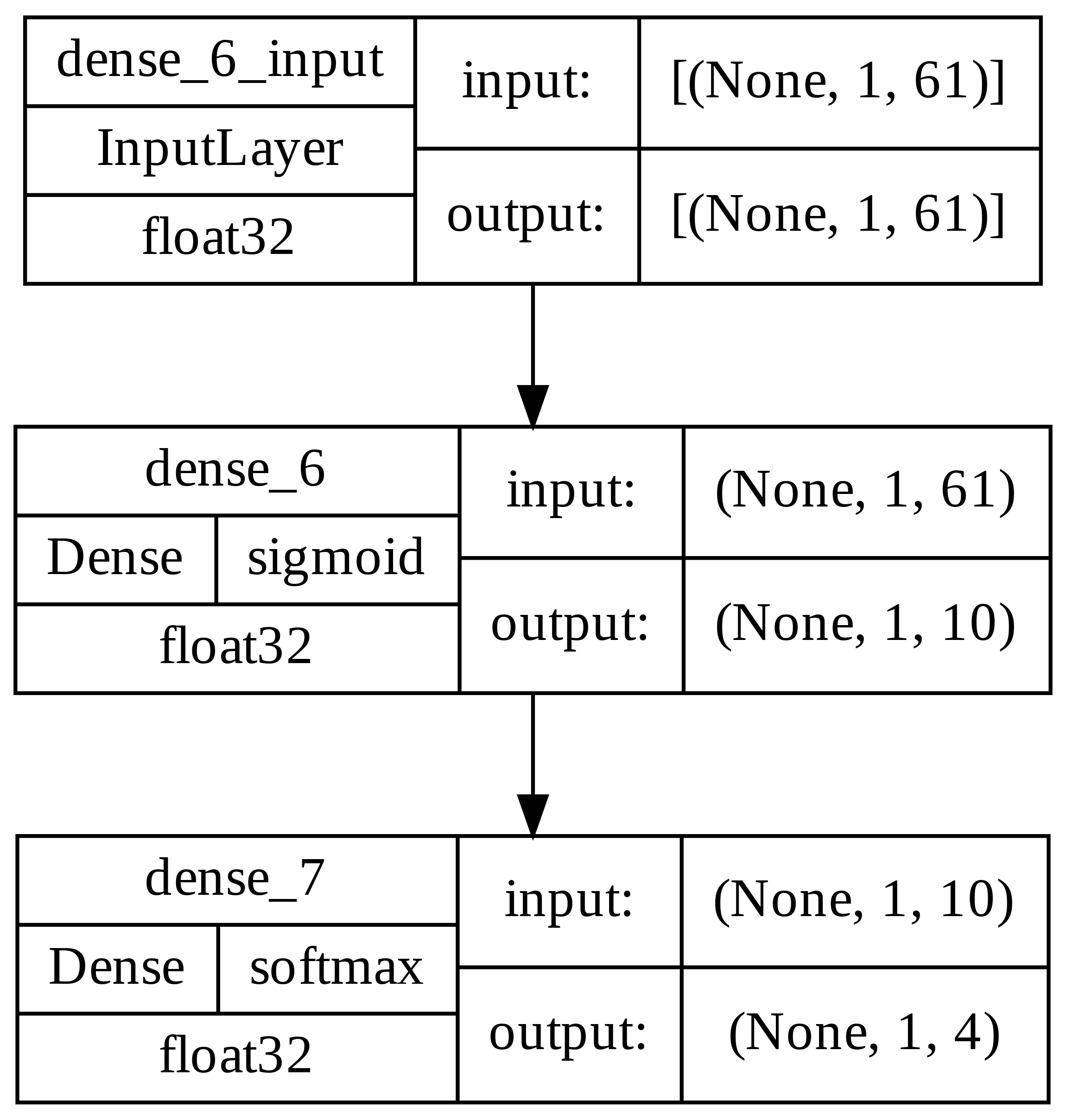}
\caption{Sigmoid-softmax based network architecture}
\label{sigmoid_softmax_structure}
\end{subfigure}
\caption{Network architecture of Sigmoid-sigmoid, ReLU-sigmoid, ReLU-softmax, and Sigmoid-softmax model}
\end{figure*}

\begin{figure*}
\centering
\begin{subfigure}[b]{0.4\textwidth}
\includegraphics[width=\textwidth]
{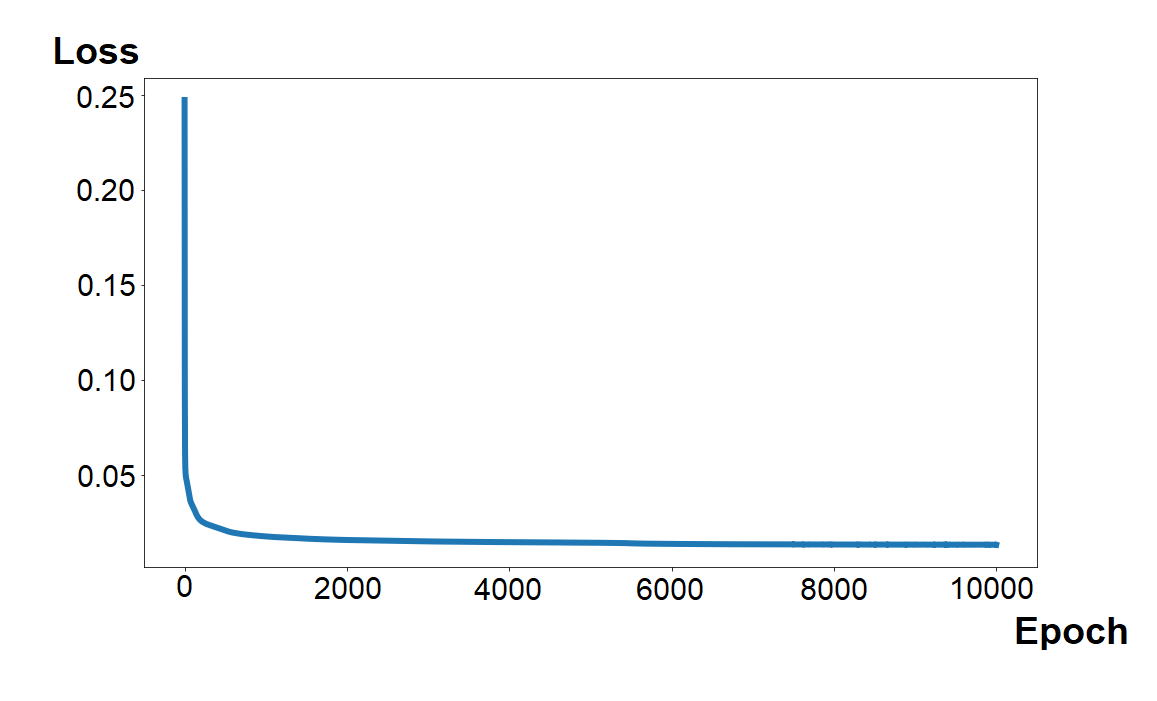}
\caption{Sigmoid-sigmoid network training convergence}
\label{sigmoid_sigmoid_loss}
\end{subfigure}
\begin{subfigure}[b]{0.4\textwidth}
\includegraphics[width=\textwidth]{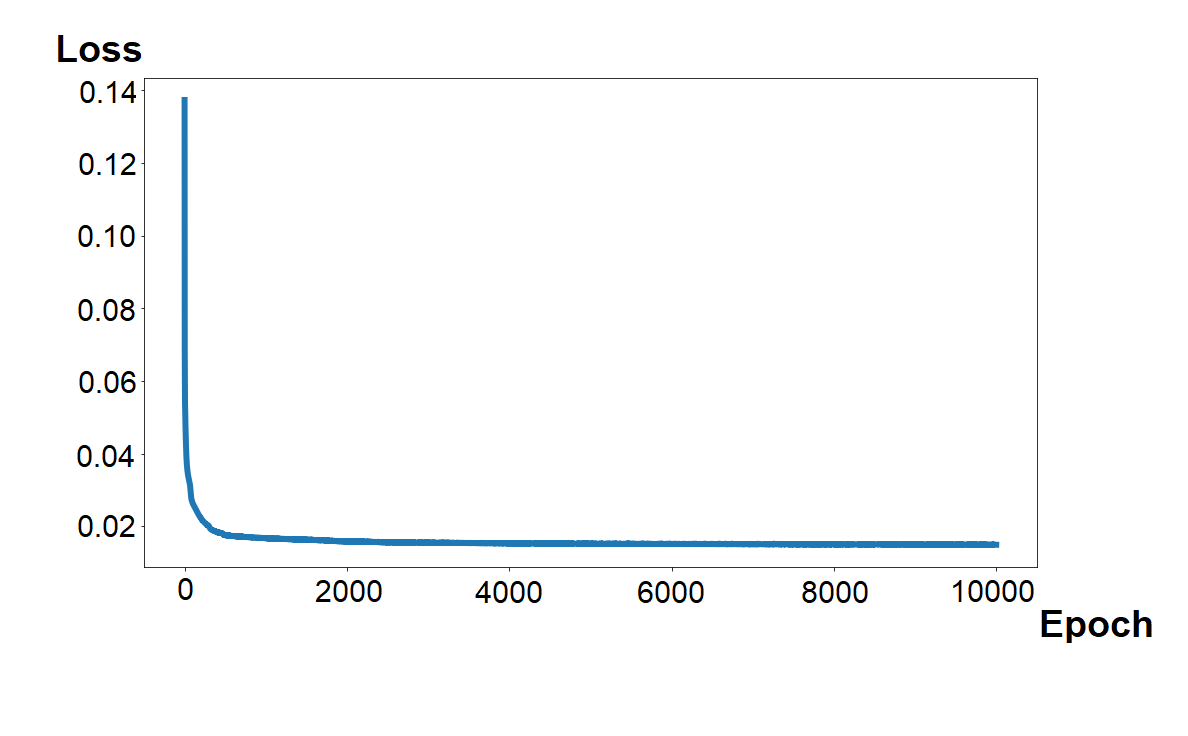}
\caption{ReLU-sigmoid network training convergence}
\label{relu_sigmoid_loss}
\end{subfigure}
\begin{subfigure}[b] {0.4\textwidth}
    \includegraphics[width=\textwidth]{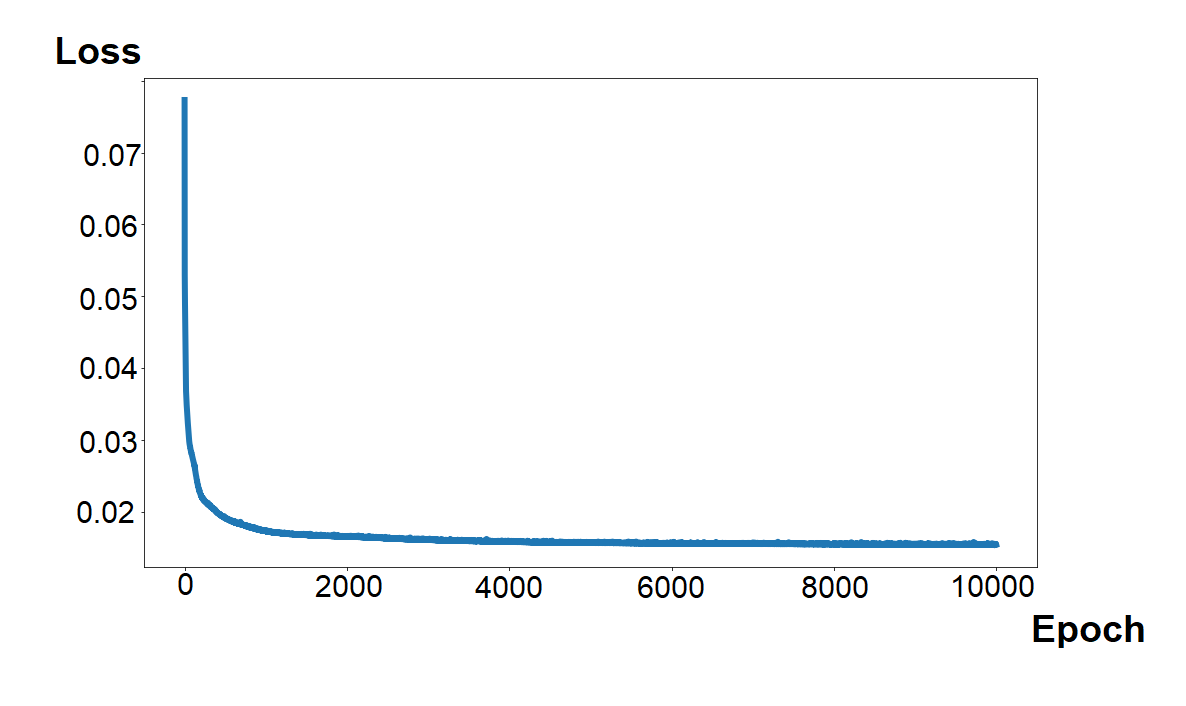}
\caption{ReLU-softmax network training convergence}
\label{relu_softmax_loss}
\end{subfigure}
\begin{subfigure}[b]{0.4\textwidth}
\includegraphics[width=\textwidth]{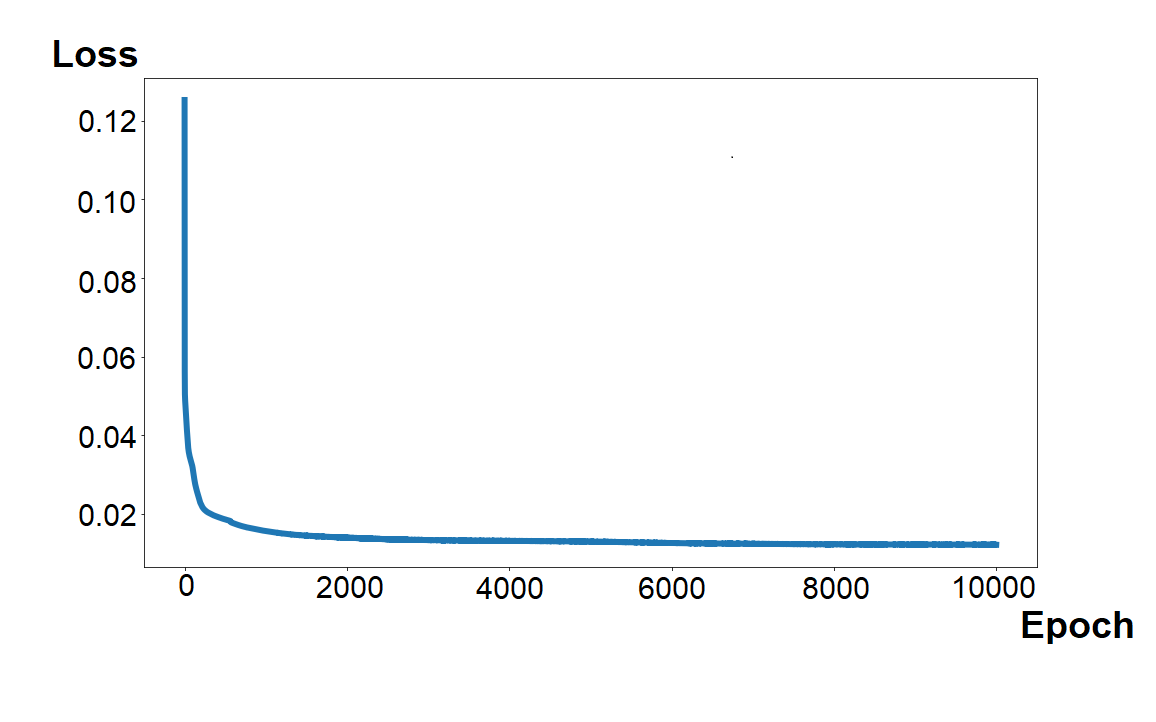}
\caption{Sigmoid-softmax network training convergence}
\label{sigmoid_softmax_loss}
\end{subfigure}
\caption{Train loss of Sigmoid-sigmoid, ReLU-sigmoid, ReLU-softmax, and Sigmoid-softmax}
\end{figure*}

The structures of our densely connected neural networks and training convergence graphs are depicted in Figure \ref{sigmoid_sigmoid_sturcture} to \ref{sigmoid_softmax_loss}. Please note that the input layer was not considered the first layer, since it had no activation function. By the first layer, we meant the first layer after the input layer. 

Hence, we used two layers, with 10 neurons in the first and four in the last. However, we tried more than one approach to find the best network. The network structures are discussed below:
\begin{itemize}

\item \textit{Sigmoid-sigmoid}:
In this network, we used the sigmoid activation function in all 14 neurons in the first and the last layer. Training achieved the highest accuracy and convergence was also the quickest.
\item \textit{ReLU-sigmoid}:
Here, we used ReLU (all 10 neurons) in the first layer and all four sigmoids in the last layer. 
\item \textit{ReLU-softmax}:
We used the softmax activation function for the last four neurons. The first 10 neurons however used ReLU.  
\item \textit{Sigmoid-softmax}:
We also tried sigmoid in the first layer and softmax activation function in the last.
\end{itemize}

Even though we are \textit{classifying} arrhythmia in this system, we opt not to choose softmax activation for a reason aside from the inferior training performance. Which is that the softmax activation function involves two consecutive loops which might delay inference time (Algorithm \ref{algo_softmax}). Besides, in an MCU, the clock is already slower, which makes saving time crucial. We chose the Sigmoid-sigmoid model for the ADM module for its best performance. We provide the variants of activation functions used from Equation \ref{sigmoid_equation} to Equation \ref{softmax_equation}. 

However, this results in the last layer allowing more than one right answer. Since the sum of the outputs of all four neurons in the last layer will not equal one as sigmoid is used. We still made this sacrifice to not miss heartbeats while processing one of them.

\begin{algorithm*}
\caption{Softmax algorithm}
\label{algo_softmax}
\begin{algorithmic}
\Require  $z$ be a tensor of rank 2 where $(rows, columns) = (1, columns)$
\Require  $n$ be the length of columns
\Ensure $sum\gets 0$

\For{\texttt{$j \gets 1$, $j\leq n$, $j \gets j+1$}}
    \State $sum \gets sum + e^{z_j}$
\EndFor

\For{\texttt{$i \gets 1$, $i\leq n$, $i \gets i+1$}}
    \State $z_i \gets \frac{e^{z_i}}{sum}$
\EndFor

\end{algorithmic}
\end{algorithm*}

\begin{equation}
\label{sigmoid_equation}
\sigma(z) = \frac{1} {1 + e^{-z}}
\end{equation}

\begin{equation}
\label{relu_equation}
Relu(z) = max(0, z)
\end{equation}

\begin{equation}
\label{softmax_equation}
\sigma(z_i) = \frac{e^{z_{i}}}{\sum_{j=1}^K e^{z_{j}}} \ \ \ for\ i=1,2,\dots,K
\end{equation}

\subsection{Error function}

We chose the mean squared error for our error/loss function as Equation \ref{MSE_single}.

\begin{equation}
E = MSE= \sum\limits_{i=1}^n \frac{(y_i - (w * x_i + b))^2}{n}
\label{MSE_single}
\end{equation}

\subsection{Optimizer}
For our optimizer algorithm, we chose the Adam optimizer developed by Kingma and Ba  \citep{adam}. The name Adam is derived from adaptive moment estimation. This optimizer uses bias-corrected moment estimates of previous timesteps' gradients to update the weight of the current step, instead of directly using the current step's gradient.

\subsection{Quantization Parameter Selection}
Quantization of our neural network parameters is necessary to fit the network within a very low memory of 2KB (see Table \ref{nano_spec}). This is why parameters trained with 32-bit floats should be converted to 8-bit integers to lose precision but gain 4$\times$ memory efficiency. 

In order to keep symmetry while quantizing our model, if the range to be quantized contains negative values, we should ensure that zero is quantized to zero.  Not only that, we noticed significant accuracy improvement across all model types tested when that was done. This is what we did in our implementation of quantization. Here, $\alpha$ and $\beta$ are the minimum and maximum range of unquantized parameters. Whereas, $\alpha_q$ and $\beta_q$ are the respective minimum and maximum of the quantized range.

Our $\alpha$ and $\beta$ range are dynamically set as the maximum quantity achieved which was $\beta=-\alpha=max=64.74442$ for the selected sigmoid-sigmoid model. However, we choose $\alpha_q = -127$ and $\beta_q = +127$.
Such choices allowed zero to be mapped to zero. We still have tested the case of zero not being mapped to zero, where $\alpha= least, \beta = max$ and $\alpha_q,\beta_q$ is kept the same (Table \ref{accuracy_drop_zero_no_zero}).

\begin{table*}[!ht]
\centering
\caption{Accuracy drop in (appx. in \%) when zero is mapped to zero vs when not while quantizing}
\label{accuracy_drop_zero_no_zero}
\begin{tabular} {|c|c|c|c|}
\hline
Quantized model& $zero_{quant}\neq 0$, $A$ & $zero_{quant} = 0$, $B$ & Drop, $B-A$\\
\hline    
Sigmoid-Sigmoid  &  80.023 & 93.74 & 13.717\\
\hline 
ReLU-Sigmoid & 89.33 & 95.40 & 6.07\\
\hline    
ReLU-Softmax &  89.33 & 94.27 & 4.94\\
\hline    
Sigmoid-Softmax & 7.13 & 95.23 & 88.1\\
\hline
\end{tabular}
\end{table*}

\subsection{Temporary Dequantization and Model Selection}
\begin{table}
\centering
\caption{Accuracy (appx. in \%) comparison between model types in default, quantized, and dequantized state }
\label{accuracy_comparison}
\begin{tabular} {|c|c|c|c|}
\hline
Model& Default & Qunatized & Dequantized\\
\hline    
Sigmoid-Sigmoid  &  97.09 & 93.74 & 96.84\\
\hline 
ReLU-Sigmoid & 96.68 & 95.40 & 96.53\\
\hline    
ReLU-Softmax & 96.63 & 94.27 & 93.33\\
\hline    
Sigmoid-Softmax & 97.21 & 95.23 & 95.51\\
\hline
\end{tabular}
\end{table}

In order to select our model for the ADM module, it is necessary that we select the state of the model in terms of quantization. In Table \ref{accuracy_comparison}, we can observe the accuracy of the model in three states, namely \textit{defualt}, \textit{quantized}, and \textit{dequantized} state. Here, \textit{default} refers to the state after the model is trained. \textit{Quantized} refers to quantizing the default model's weight/bias to an 8-bit integer from -127 to +127 in this case. \textit{Dequantized} state means, scaling the weights/biases back to default. However, the weights/biases may not be equal to the default since quantizing involves rounding (see Equation \ref{eqn_dequantize}). However, they will be approximate. Table \ref{weights_quantized_dequantized} includes some weights and biases in their default, quantized, and dequantized state along with their corresponding accuracy. We see that in the dequantized state, they do not recover their exact precision but are close. 

\begin{equation}
\label{eqn_dequantize}
x = s (x_q + z) 
\end{equation}

\begin{table*}
\centering
\caption{Default, quantized, dequantized state of weight-bias for each layer and corresponding accuracy (appx in \%) for the sigmoid-sigmoid model}
\label{weights_quantized_dequantized}
\begin{tabular} {|c|c|c|c|c|c|}
\hline
Weight/bias type&  {Layer 1 weight} & {Layer 1 bias}&  {Layer 2 weight} & {Layer 2  bias} & {Accuracy}\\
\hline    
Default & -56.74 & -1.58 & 14.58 & -17.0 & 97.09\\
\hline 
Quantized & -111 & -3 & 29 & -33 & 93.74\\
\hline    
Dequantized & -56.59 & -1.53 & 14.78 & -16.82 & 96.84\\
\hline
\end{tabular}
\end{table*}

Even though it is not possible to pack one of the default models to the MCU, since their size exceeds 2KB, we can pack the quantized and the dequantized weights/biases. Using quantized weights/biases will be the easiest since they will take the least space and work seamlessly. However, packing dequantized state is also possible through \textit{temporary dequantization}. This means packing the parameters as quantized 8-bit integers. However, reverting back to 32-bit floats at inference time. This will not overload the memory since only one parameter will dequantize at a time. That is why we have decided to use the dequantized sigmoid-sigmoid model with the temporary dequantization technique since it has the highest accuracy among the dequantized and quantized models.

In this technique, weights can stay as 8-bit integers in memory (as quantized integers). However, when that specific weight/bias is used for calculation, it can be copied, dequantized, used, and released from memory. Hence, we have packed the sigmoid-sigmoid quantized weights/biases along with the calculated scale and zero point. Then when the time came for inference using any of the above weights, we dequantized only that specific weight/bias and caused inference. That way at inference time, we are only using three bytes of additional memory. Which is tolerable, provided the increase in performance is achieved.  Algorithm \ref{algo_dequantize} shows the use of temporary dequantization as implemented in the MCU.  

However, the dequantization method as shown in Algorithm \mbox{\ref{algo_dequantize}} is slower than using quantized parameters. In Table \mbox{\ref{dequant_time_cost}} we observe that for the same asymptotic time costs, the dequantized time costs are $28.74$\% and $14.67$\% slower than the quantized time costs for neural layer 1 and 2 respectively. To measure the on-device time costs in Table \mbox{\ref{dequant_time_cost}}, we have run our model on Nano and sampled the time cost for n=100 times for mean ($\mu$) and standard deviation ($\sigma$) calculation.

\begin{table*}
\centering
\caption{Algorithm \mbox{\ref{algo_dequantize}} time cost, dequantized vs quantized}
\label{dequant_time_cost}
\begin{tabular}{|c|c|c|c|c|c|c|c|c|c|} \hline 

\multicolumn{1}{|p{1cm}|}{\centering Neural layer no.} & \multicolumn{1}{p{1cm}|}{ \centering Neuron} &\multicolumn{1}{p{1cm}|}{\centering Input row, $m$} & \multicolumn{1}{p{2cm}|}{\centering Input column \& kernel row, $n$} & {\centering Kernel column, $p$}& \multicolumn{1}{p{2cm}|}{\centering Asymptotic cost, $O(m.n.p+p)$}& \multicolumn{2}{p{2cm}|}{\centering Dequantized cost, on device (ms), samples$=$100}  & \multicolumn{2}{p{2cm}|}{\centering Quantized cost, on device (ms), samples$=$100} \\ 
\hline
{} & {} & {} & {} & {} &{}& {$\mu$} & {$\sigma$} & {$\mu$} & {$\sigma$} \\
\hline
1 & 10 & 1 & 61 & 10 & $O(620)$ & 17.880 & 0.325 & 12.740 & 0.439\\ 
\hline 
2 & 4 & 1 & 10 & 4 & $O(44)$ & 2.250 & 0.433 & 1.920 & 0.271\\ 
\hline
\end{tabular}
\end{table*}

\begin{algorithm*}[ht]
\caption{Application of temporary dequantization}
\label{algo_dequantize}
\begin{algorithmic}
\Require  $input\;tensor$ be a tensor of rank 2
\Require  $kernel\;tensor$ be a tensor of rank 2
\Ensure $columns\;of\;input\;tensor = rows\;of\;kernel\;tensor$
\Procedure{matmul}{$input\;tensor$, $kernel\;tensor$}
\For{\texttt{$i \gets 1$, $i\leq rows\;of\;input\;tensor$, $i \gets i+1$}}
    \State $result_{ij} \gets 0$
    \For{\texttt{$j \gets 1$, $j\leq columns\;of\;kernel\;tensor)$, $j \gets j+1$}}
    \For{\texttt{$k \gets 1$, $k\leq columns\;of\;input\;tensor$, $k \gets k+1$}}
        \State $result_{ij} \gets result_{ij} + dequantize(input_{ik} \cdot kernel_{kj})$ 
        \Comment{Dequantization applied here}
    \EndFor
\EndFor
\EndFor \\
\Return $result$
\EndProcedure

\Procedure{NeuralLayer}{$input\;tensor$, $kernel\;tensor$, $bias$}
    \State $result \gets matmul(input\;tensor,\;kernel\;tensor)$
    \For{\texttt{$i \gets 1$, $i\leq columns\;of\;bias\;tensor$, $i \gets i+1$}}
        \State $result_{ij} \gets \sigma(result_{ij} + dequantize(bias_i))$ 
        \Comment{Dequantization applied here}
    \EndFor \\
    \Return $result$
\EndProcedure

\end{algorithmic}
\end{algorithm*}


\section{Experimental Evaluation}
\label{sec_exp}
This Section details our training setup, results, and discussion of the densely connected neural network or ADM discussed in the previous section. 

\subsection{Experimental Setting}
In this subsection, we discuss the training hardware specification of the arrhythmia detection module and the dataset used. 

\textit{Training Specification}: In order to train our densely connected neural network model for classifying arrhythmia we have used Google Colab \citep{colab}. The specification of the freemium version of Google Colab is provided in Table \ref{table_exp_set}. Our models were trained for 10,000 epochs and the learning rate was $0.001$. We have used batched training of 1024 samples per epoch.

\begin{table}[!ht]
\centering
\caption{Computer specification for our neural network model training}
\label{table_exp_set}
\begin{tabular} {|c|c|}
\hline
\text{Hardware type} & \text{Description} \\
\hline
\text{System RAM} &  12.7 GB  \\
\hline 
\text{CPU} & Intel(R) Xeon(R) CPU @ 2.20 GHz\\
\hline 
\text{GPU} & NVIDIA Tesla T4 \\
\hline    
\text{Disk space} & 78.2 GB\\
\hline
\end{tabular}
\end{table}

\textit{MIT-BIH Dataset \citep{mitbih_dataset}}: The research on arrhythmia analysis and related topics has been supported since 1975 by labs at MIT and Boston's Beth Israel Hospital (now the Beth Israel Deaconess Medical Center). The MIT-BIH Arrhythmia Database, which was finished and started disseminating in 1980, was one of the first significant outcomes of that work. The database, which was used in about 500 locations around the world, was the first generally accessible collection of standard test materials for arrhythmia detector validation. In addition to the assessment of arrhythmia detection, it was also used for fundamental study into cardiac dynamics. Initially, the dataset was delivered on a quarter-inch IRIG-format FM analog tape and 9-track half-inch digital tape at 800 and 1600 bpi. A CD-ROM version of the database was published in August 1989. From 1975 to 1979, 48 half-hour snippets of two-channel ECG recordings from 47 people were kept in the database. 23 recordings were chosen from 4000 24-hour ECG recordings where the inpatient-outpatient ratio is 60-40. The rest of the recordings (25) were selected from the same set in order to include less common but clinically significant arrhythmia.

The recordings were digitized over a 10 mV range at 360 samples per second per channel with an 11-bit resolution. Two cardiologists separately annotated each record; differences were settled to produce the computer-readable reference annotations for each beat, which are supplied with the database and total over 110,000 annotations. Since the launch of PhysioNet \mbox{\citep{physionet}} in September 1999, around half of this database—25 of 48 complete records, and reference annotation files for all 48 entries—has been freely accessible through the website of PhysioNet. In February 2005, the final 23 signal files that had previously only been accessible via the MIT-BIH Arrhythmia Database CD-ROM were posted. The entire dataset is now accessible \footnote{\href{https://physionet.org/content/mitdb/1.0.0/} physionet.org/content/mitdb/1.0.0/}.

\begin{table}
\centering
\caption{Train-test split of the used dataset}
\label{train_test_split}
\begin{tabular}{|c|c|c|c|c|c|}
\hline
Samples & N & S & V & F & Total \\
\hline
Training  & 60723 & 1863 & 4848 & 538 & 67972\\
\hline
Testing   & 29908 & 918 &  2388 &  265 & 33479\\ 
\hline
Total   & 90631 & 2781 & 7236 & 803 & 101451\\ 
\hline
\end{tabular}
\end{table}

From this dataset, we used four types of beats as suggested by AAMI \citep{aami} as shown in Table \ref{arrhythmia_classification}. The train test split of the dataset for training ADM is provided in Table \ref{train_test_split}.

Besides the used dataset, numerous similar datasets such as the CINC 2020 Challenge database \mbox{\citep{alday2020classification}}, the Chapman ECG database \mbox{\citep{zheng202012}}, and the PTB Diagnostic ECG Database \citep{ptb} are available. However, for arrhythmia classification we found the MIT-BIH Arrhythmia database to be the most useful resource.

\subsection{Experimental metrics}
To evaluate our neural network model, we use precision, recall, and F1 score along with accuracy as metrics since accuracy alone can be misleading. The equations for the metrics are provided from Equation \mbox{\ref{eqn_precision}} through \mbox{\ref{eqn_f1}}. Table \mbox{\ref{true_false_posneg}} contains the explanation for TP and FP used in the equations. To understand them, an explanation of some concepts is necessary -- 
\begin{itemize}
\item \textit{True positive}: Number of samples detected as positive and are also positive.
\item \textit{False positive}: Number of samples detected as positive but are negative.
\item \textit{True negative}: Number of samples detected as negative and are also negative. 
\item \textit{False negative}: Number of samples detected as negative but are positive.      
\end{itemize}

\begin{table}[!ht]
\centering
\caption{True vs False positives and negatives}
\label{true_false_posneg}
\begin{tabular}{|c|c|c|c|c|c|c|c|}
\hline
{} & Ground Truth & Predicted \\
\hline
TP & Positive & Positive \\
\hline
FP & Negative & Positive \\
\hline
TN & Negative & Negative \\
\hline
FN & Positive & Negative \\
\hline
\end{tabular}
\end{table}

Here, precision is a ratio of detected true positives for a specific class in a classification problem with a total number of testing samples detected as positive (both true and false). The recall is the ratio of true positives but with the total number of samples for that class. A fall in precision means an increase in false positives. On the other hand, a fall in recall means a rise in false negatives. The rise consequently means the opposite for both. For which high recall and precision are always desirable. However, in some cases, only one of them being higher than the other also works as long as both are high enough. 

The F1 score is the harmonic mean of precision and recall. A harmonic mean for two numbers is only high when all of its members are high. Therefore using the F1 score, we can know whether both the recall and precision are high or not. That way we can use a single metric of performance instead of two. The macro average is the average of any of the classification scores such as precision, recall, or f1 (Equation \mbox{\ref{eqn_macro_avg}}). However, in the weighted average, each score is weighted by the no. of times it appears in the dataset (Equation \mbox{\ref{eqn_weighted_avg}}). That way the mean score accounts for the success/failure it had for each class. 

\begin{equation}
\label{eqn_precision}
Precision = \frac{TP}{TP + FP}
\end{equation}
\begin{equation}
\label{eqn_recall}
Recall = \frac{TP}{TP + FN}
\end{equation}
\begin{equation}
\label{eqn_f1}
F1 = \frac{2 * Precision * Recall}{Precision + Recall}
\end{equation}
\begin{equation}
\label{eqn_macro_avg}
Macro\;average = \frac{\sum\limits_{i=1}^{Number\;of\;classes} Score_i}{Number\;of\;classes} 
\end{equation}
\begin{equation}
\label{eqn_weighted_avg}
Weighted\;average = \frac{\sum\limits_{i=1}^{C} Samples_i * Score_i}{\sum\limits_{i=1}^{C} Samples_i} 
\end{equation}

\subsection{Experimental Results and Findings}
In this Section, we show the performance of the dequantized sigmoid-sigmoid model since it has achieved the highest accuracy with the ability to be compressed to fit in MCU's memory. The performance details are provided in Table \ref{perf_selected_model}. The scores are based on the test dataset mentioned in Table \mbox{\ref{train_test_split}}. Consecutively, we provided the second and the third best model in the same table. 

\begin{table*}
\caption{Performance of best three models. The sigmoid-sigmoid dequantized model was selected among the three.}
\label{perf_selected_model}
\begin{tabular}{|c|c|c|c|c|c|c|c|c|}
\hline
{} & {} &             N &           S &            V &           F &       Weighted avg & Macro avg  & Accuracy \\ 
\cline{2-9}
\multirow{4}{8em}{Sigmoid-sigmoid Dequantized *} & Precision &      0.975297 &    0.783034 &     0.928136 &    0.937500 &    0.966362   & 0.905992   & {}\\ 
\cline{2-8}
{} & Recall    &      0.992711 &    0.522876 &     0.892379 &    0.452830 &    0.968398   & 0.715199   &  0.968398 \\
\cline{2-8}
{} & F1-score  &      0.983927 &    0.627041 &     0.909906 &    0.610687 &    0.965907   & 0.782890   & {} \\ 
\hline

\multirow{4}{8em}{Relu-softmax Dequantized} & 
Precision &      0.970984 &    0.817269 &     0.932057 &    0.780000  &     0.962481    &  0.875077   &  {} \\ 
\cline{2-8}
{} & Recall    &      0.994684 &    0.443355 &     0.855946 &    0.441509  &    0.965292  &    0.683874    & 0.965292\\ 
\cline{2-8}
{} & F1-score  &      0.982691 &    0.574859 &     0.892382 &    0.563855 &     0.961751  &  0.753447    &  {}\\ 
\hline

\multirow{4}{8em}{Sigmoid-sigmoid Quantized} & 
Precision &      0.980569 &    0.369759 &     0.819113 &    0.558065 &    0.948960    &   0.681876   & {}\\ 
\cline{2-8}
{} & Recall    &      0.953324 &    0.586057 &     0.904523 &    0.652830  & 0.937394 &  0.774183       & 0.937394\\ 
\cline{2-8}
{} & F1-score  &      0.966754 &    0.453434 &     0.859701 &    0.601739 &  0.942154 &  0.720407      & {}\\ 
\hline

Testing Samples   & {} & 29908 &  918 &  2388 &  265 &  33479  & {} &  {}\\ 
\hline
\end{tabular}
\end{table*}

\begin{table*}
\small
\centering
\caption{Performance comparison with the best arrhythmia classification models studied}
\label{perf_comparision_studies}
\begin{tabular}{|c|c|c|c|}
\hline
\multicolumn{1}{|p{1cm}}{\centering Model} & \multicolumn{1}{|p{2cm}|}{\centering Faraone et al.,2020 \citep{faraone}} & \multicolumn{1}{p{2cm}|}{\centering Scirè et al.,2019  \citep{scire}}& \multicolumn{1}{p{2.3cm}|}{\centering \hl{Sigmoid-sigmoid network (Proposed)}}\\
\hline
FLOPs (MOps) & 3.221  & - & 0.001314 \\
\hline
Memory (KB) &  6.8  & -  &1.267 \\
\hline 
Precision & 0.795 & 0.805 & 0.905 \\ 
\hline
Recall & 0.776 & 0.891 & 0.715  \\ 
\hline
F1-score & 0.780 & 0.845 & 0.782   \\ 
\hline
Accuracy &  0.854 & 0.968 & 0.968 \\
\hline
\end{tabular}
\end{table*}

We observe in Table \ref{perf_selected_model} that the model has nearly perfect precision and an even recall for type N arrhythmia. Which provides a great F1 score. The second-best performance was achieved by type V arrhythmia. However, the scores for types S and F are not satisfactory. Although type F has a higher precision, the recall is less than 50\%. On the other hand, the recall score for S type is slightly higher than type F. However both have close F1 scores. Which means the overall performance type S and F is the same. The reason for this is quite apparent since the training samples of S and F are close to 2000 and 500 respectively (please see Table \ref{train_test_split}). However, type N has nearly 60,000 samples and V has almost 5000 samples which is significantly higher than S and F. Moreover, the weighted average for each score is very high, implying the success of classifying the N and V type arrhythmia correctly. However, in the case of the macro average which implies as said before, the model would have performed better if a more balanced dataset was provided.

To verify this conjecture we have trained the model by randomly sampling 500 samples per class of arrhythmia from the train set of Table \mbox{\ref{train_test_split}}. We set 500 to be the sampling number since the arrhythmia F has 538 samples in the train set which is the lowest number of samples and possibly the root cause for the worst performance of type F arrhythmia classification. We observe in Table \mbox{\ref{model_less_data}}, the performance before quantization-dequantization to be significantly poorer than the dequantized sigmoid-sigmoid model of Table \mbox{\ref{perf_selected_model}}. This means after quantization and subsequent dequantization the performance ought to get worse. This proves that if S and F had more training examples in the dataset, the performance could be better. Likewise, the performance of all classes of arrhythmia is worse due to a lack of enough samples.

Moreover, it is also of immense significance to check for overfitting in the network. In Table \mbox{\ref{model_train_test_perf}} we observe that the training and testing accuracy are both approximately 97\%. On the other hand, the f1-score of both train and test datasets are approximately 83\% and 82\% respectively which are also close. The scores have been taken before, the sigmoid-sigmoid model has been quantized and dequantized, hence the scores are better than Table \mbox{\ref{perf_selected_model}}. Since the training and testing scores are close, this proves the model is not suffering from overfitting.

\begin{table}
\centering
\caption{Performance of the sigmoid-sigmoid network trained with only 500 samples per class}
\label{model_less_data}
\begin{tabular}{|c|c|c|}
\hline
Dataset & Train & Test \\
\hline
Samples & 500 & 33479\\
\hline
Accuracy & 0.8080 & 0.71767\\
\hline 
F1 Score (macro avg.) & 0.80833 & 0.46744\\
\hline
\end{tabular}
\end{table}

\begin{table}[h]
\centering
\caption{Train-test performance comparison of the default sigmoid-sigmoid model}
\label{model_train_test_perf}
\begin{tabular}{|c|c|c|}
\hline
Dataset & Train & Test \\
\hline
Samples & 67972 & 33479\\
\hline
Accuracy & 0.9720 & 0.9709\\
\hline 
F1 Score (macro avg.) & 0.83407 & 0.81960\\
\hline
\end{tabular}
\end{table}

Furthermore, analyzing FLOPs (Floating Point Operations) we observe our operation count is 1314 operations per inference. Also, the model weights and biases consume 664 bytes of memory. But we at least need to keep 150 samples of ECG data in memory for the  HDM module which is eventually preprocessed to 61 samples that the ADM module intakes for inference. Each of the 150 samples occupies 4 bytes, whereas the model weight/bias occupies 1 byte each. Also for temporary dequantization, we require an additional 3 bytes at any moment in time. That amounts to 1267 bytes or 1.267 KB of memory. However, this calculation is an approximation of the \textit{unreleasable} memory. In practice, we had to allocate and deallocate some more memory. The calculations are provided in Table \ref{flops_mem_per_layer} and \ref{sram}.

\begin{table}
\centering
\caption{FLOPs \& memory breakdown per layer}
\label{flops_mem_per_layer}
\begin{tabular}{|c|c|c|}
\hline
Layer & FLOPs & Memory\\
\hline
1 & 1230 & 620 \\
\hline
2 & 84 & 44 \\
\hline 
Total & 1314 & 664\\ 
\hline
\end{tabular}
\end{table}

\begin{table}
\centering
\caption{System SRAM consumption breakdown in MCU}
\label{sram}
\begin{tabular}{|c|c|}
\hline
{} & Memory (bytes) \\
\hline
Model size & $664 + 3 = 667$ \\
\hline
ECG buffer & $150 \cdot 4 = 600$ \\
\hline 
Total & 1267 \\ 
\hline
\end{tabular}
\end{table}
Previously, we have discussed that Faraone et al. \citep{faraone} and Scire et al. \citep{scire} achieved better performance than all other compared studies. Therefore, Table \ref{perf_comparision_studies} contains the performance of their arrhythmia classification model vs ours. This is to be noted that the performance metrics of Table \mbox{\ref{perf_comparision_studies}} (accuracy, precision, recall, and f1-score) are from Table \mbox{\ref{perf_selected_model}} and based on the test set of Table \mbox{\ref{train_test_split}}. In Table \mbox{\ref{perf_comparision_studies}}, we notice that the FLOPs and memory calculation of Scire et al. are absent. The reason is that we could not find explicit data or the number of weights or biases used by them per layer in the study. However, compared with the available data we notice that our model consumes more than 5 times less memory and 2500 times fewer operations than Faraone et. al, where our performance exceeds theirs by 0.2\% in terms of F1 score and 11\% in terms of accuracy. However, we underperformed compared to Scire et al. by 6.3\% in terms of F1 score where our accuracy was equal.

In addition, this study shows that neural network models using fewer weights/biases and primitive activations like sigmoid can outperform or come close to modern networks with a high parameter count using dropout and Convolution layers. Moreover, we noticed the relu-sigmoid dequantized and sigmoid-sigmoid quantized models to provide the second and third-best performances whose F1 scores were 0.753 and 0.720 respectively whereas the selected sigmoid-sigmoid dequantized model has 0.782 in Table \ref{perf_selected_model}. Using different random seeds, one might get a higher performance for them since those models have very close scores. This might make temporary dequantization redundant (if the sigmoid-sigmoid quantized model gets a better f1 score) and ensure faster inference.

\subsection{Further Optimization}
Besides quantization, there are numerous neural network compression techniques available that reduce the size of the network without compromising performance if done right. However, in the case of a minimal network like ours with only 664 parameters, it might prove difficult to compress any further without losing performance. To verify our hypothesis, we perform several experiments on our \mbox{\textit{sigmoid-softmax}} model as our base model. The experimental setup and discussion of the results in Table \mbox{\ref{model_optimization}} are described in the Subsections \mbox{\ref{model_pruning}} through \mbox{\ref{weights_only}}.
 
\subsubsection{Parameter Pruning}
\label{model_pruning}
Parameter pruning involves setting some of the parameters to zero based on thresholds. The thresholds are calculated per layer based on the parameters L0, L1, or L2 regularization. In this experiment, we pruned the weights based on L0 regularization and then retrained our model. Specifically, all parameters per layer and per parameter type, are set to zero if their absolute value is less than the $50^{\text{th}}$ percentile absolute value of that layer and parameter type. For instance, the thresholds of layer 1 weights are set, based solely on the least  $50^{\text{th}}$ percentile absolute value of the layer 1 weights. Similarly, the thresholds of layer 1 bias are set based on the bias of the same layer. The same goes for layer 2. This was done since bias and weight regulate two different aspects of a neuron, hence one should not be considered to regulate the other. After pruning, we retrained our model without iterative pruning with a learning rate $1/100$ of our original learning rate \citep{han2015pruning}. However, we can see there is a significant performance decrease (1.3$\times$) in the F1 score compared to the base model in Table \mbox{\ref{model_optimization}}.

\subsubsection{Knowledge Distillation}
\label{distillation}
Knowledge distillation is the training of a smaller neural network with the soft targets produced by a larger model where the loss function is a sum of two losses. One is the loss between soft targets produced by the larger and smaller models. The soft targets are scaled/divided by a temperature value. The other loss is the loss between hard targets from the dataset and soft targets produced by the model. Typically, multipliers determine which of the two losses should have more effect \citep{hinton2015distilling}. The larger model and the smaller model are typically called teacher and student respectively. We trained a smaller model of a total of 8 neurons in a 4-4 structure as our student model with our base model (sigmoid-softmax) as the teacher model with 14 neurons in a 10-4 structure. We set the temperature to 10 for training. The soft and hard target loss multipliers are 0.9 and 0.1 respectively. We used categorical crossentropy for our hard-target loss and Kullback–Leibler divergence loss for soft target loss. We can observe the distilled model to have a similar performance decrease as \mbox{\ref{model_pruning}} in Table \mbox{\ref{model_optimization}}.

\subsubsection{Using Only Weights as Parameters}
\label{weights_only}
We discarded the biases of our base model and trained it from scratch without changing the structure of our training. However, with the same training configuration, the model achieved slightly inferior performance in both accuracy and f1 score as observed in Table \mbox{\ref{model_optimization}}.

Since the above optimization techniques show inferior performance, we have decided not to apply them to our final model. In the field of medical diagnosis, developing tools that provide the most accurate measurement is a must. With a few exceptions, performance should not be compromised to attain efficiency. However, we had to deploy quantization to ensure that our model ($2.59$ KB) fits within the SRAM ($2$ KB) of our selected MCU. Hence, the performance decrease due to quantization was unavoidable in our case. Therefore, we have avoided the above optimization techniques to ensure the best performance.

\begin{table*}
\centering
\caption{Performance comparison of base sigmoid-softmax, pruned, distilled, and weights only model}
\label{model_optimization}
\begin{tabular}{|c|c|c|c|c|}
\hline
Model & Base & Pruned & Distilled & Weights only\\
\hline
Accuracy & 0.9689 & 0.9536 & 0.9578 & 0.9679 \\
\hline 
F1 Score (macro avg.) & 0.8102 &  0.6157 & 0.6018 & 0.7901 \\
\hline
\end{tabular}
\end{table*}

\section{Conclusion and Future Work}
\label{conclusion}
\hl{In this study, an efficient fully connected neural network has been developed for detecting arrhythmia or anomalous heartbeats. Remarkably, this network is designed to operate on extremely resource-constrained devices, such as microcontroller units (MCUs). 
Nonetheless, to guarantee its practical applicability, it was imperative to create the entire end-to-end ECG monitoring system. 
Undoubtedly, an ECG system equipped with a low-compute processor will not only be more cost-effective but will also consume less power}. Using such ECG monitoring systems could reduce the costs of treating CVD patients worldwide. 
Future work with Edge Computing has a great prospect for \hl{such applications}. \hl{A high-compute machine at the edge could process the classified arrhythmia collected from low-compute systems to predict heart attacks or other anomalies before they occur}. 

Furthermore, Cloud Computing is prospective for the same application but on a much larger scale. \hl{In conclusion, it can be said that fitting neural network models in low compute systems, can promise an easy and efficient living for all}. For the highly expensive biomedical sector, the statement is more promising and truer than ever. If more biomedical devices like ECG \hl{follow} suit to develop with low-compute machines, millions more lives could be easily saved.

\balance
\bibliographystyle{apacite}
\bibliography{cas-refs}

\end{document}